\documentclass[a4paper,10pt]{amsart}
 \usepackage{amsfonts,mathrsfs}
 \usepackage{amsrefs}
 \usepackage{amsthm}
 \usepackage{amssymb}
 \usepackage{enumerate}
 \usepackage{graphicx}
 \usepackage{tikz-cd}
 \usepackage{cleveref}
 \usepackage{cancel}
\allowdisplaybreaks

\theoremstyle{definition}
\newtheorem{Def}{Definition}[section]

\newtheorem{rem}[Def]{Remark}

\theoremstyle{plain}

\newtheorem*{thm*}{Theorem}

\newtheorem*{cor*}{Corollary}

\newtheorem*{con*}{Conjecture}
\newtheorem*{frag*}{Question}
\newtheorem*{verm*}{Vermutung}

\newtheorem{theorem}{Theorem}
\newtheorem{proposition}{Proposition}

\newtheorem{corollary}{Corollary}

\theoremstyle{definition}
\newtheorem{example}{Example}
\usepackage{url}
\usepackage{kbordermatrix}
\theoremstyle{definition}
\newtheorem{remark}{Remark}
\usepackage{enumitem}
\numberwithin{equation}{section}

\makeatletter
\newcommand{\mylabel}[2]{#2\def\@currentlabel{#2}\label{#1}}
\makeatother

\usepackage[section]{algorithm}
\usepackage{algpseudocode}

\usepackage[a4paper,margin=2.5cm]{geometry}
\newcommand{\dd}{{\rm d}}
\newcommand{\mj}{\mathbb{J}}

\newcommand{\me}{\mathbb {E}}
\newcommand{\x}{\underline{x}}
\newcommand{\uu}{\underline{u}}

\newcommand{\RR}{\mathbb R}
\newcommand{\NN}{\mathbb N}

\newcommand{\calD}{\mathcal D}

\newcommand{\calL}{\mathcal L}

\renewcommand{\kbldelim}{(}
\renewcommand{\kbrdelim}{)}
\renewcommand{\kbrowstyle}{\displaystyle}
\renewcommand{\kbcolstyle}{\displaystyle}
\newcommand{\CB}[1]{\begin{color}{blue}#1\end{color}}

\DeclareMathOperator{\opt}{\ast}

\newcommand{\cB}{{\mathcal B}}

\newcommand{\cD}{{\mathcal D}}

\usepackage{mathrsfs}
\usepackage{bbm}
\newcommand{\Int}{\operatorname{int}}

\title[Extreme mass distributions for quasi-copulas]{Extreme mass distributions for quasi-copulas}
\author{Matja\v z Omladi\v c${}^{1}$}
\address{Matja\v z Omladi\v c,
Faculty of Mathematics and Physics, University of Ljubljana  \& Institute of Mathematics, Physics and Mechanics, Ljubljana, Slovenia.} 
\email{matjaz@omladic.net}\thanks{${}^1$Supported by the ARIS (Slovenian Research and Innovation Agency)
research core funding No.\ P1-0448.}
\author{Martin Vuk}
\address{Martin Vuk, 
Faculty of Computer and Information Science, University of Ljubljana.}
\email{martin.vuk@fri.uni-lj.si}
\author{Alja\v z Zalar${}^{3}$}
\address{Alja\v z Zalar, 
Faculty of Computer and Information Science, University of Ljubljana  \& 
Faculty of Mathematics and Physics, University of Ljubljana  \&
Institute of Mathematics, Physics and Mechanics, Ljubljana, Slovenia.}
\email{aljaz.zalar@fri.uni-lj.si}
\thanks{${}^3$Supported by the ARIS (Slovenian Research and Innovation Agency)
research core funding No.\ P1-0288 and grants J1-50002, J1-60011.}

\newcommand{\comment}[1]{}
\newcommand{\mi}{\mathbb{I}}
\newcommand{\sign}{\mathrm{sign}}

\begin{document}

\subjclass[2020]{Primary 62H05, 60A99; Secondary 60E05, 26B35.}

\date{\today}
\keywords{mass distribution, $d$-quasi-copula, volume, Lipschitz condition, bounds}

\begin{abstract}
  The recent survey \cite{ARIASGARCIA20201} nicknamed ``Hitchhiker's Guide'' has raised the rating of quasi-copula problems in the dependence modeling community in spite of the lack of statistical interpretation of quasi-copulas. 
  In our previous work we addressed the question of extreme values of the mass distribution associated with a mutidimensional quasi--copulas. 
Using linear programming approach we were able to settle
 \cite[Open Problem 5]{ARIASGARCIA20201} up to $d=17$ and disprove a recent conjecture from \cite{UF23} on solution to that problem. In this note we use an analytical approach to provide a complete answer to the original question.
\end{abstract}

\maketitle

\section{Introduction}
\label{sec:intro}


Copula is simply a multivariate distribution with uniform margins, but when we insert arbitrary univariate distributions as margins  into it, we can get any multivariate distribution. This seminal 1959 result of Sklar \cite{Sklar1959229} has made them the most important tool of dependence modeling \cite{Nel05,durante2015principles}. On the other hand, the statistical interpretation of quasi-copulas is problematic since they may have negative volumes locally. They come naturally as (pointwise) lower and upper bounds of sets of copulas. 
The lower Fr\'{e}chet-Hoeffding bound $W$, say, the pointwise infimum of all $d$-variate copulas, $d\ge2$, is in general a quasi-copula, unless $d=2$, while the upper bound $M$, the pointwise supremum of all $d$-variate copulas, is always a copula. Quasi-copulas were first introduced in 1993 by Alsina, Nelsen, and Schweizer \cite{Als93}, but the dependence modeling community may have been somehow withheld from wider use of them perhaps due to their deficiency described above. 

The new era for this notion starts in 2020 with the seminal paper  \cite{ARIASGARCIA20201} by Arias-Garc\'{i}a, Mesiar, and De Baets, listing the most important results and open problems in the area. Some vivid activity followed this paper.  The six open problems listed there are sometimes nicknamed ``hitchhiker’s problems''. The first one solved seems to have been hitchhiker's problem \#6, Omladi\v c and Stopar \cite{OmSt1} in 2020; the second one was hitchhiker's problem \#3, Omladi\v c and Stopar \cite{OmSt5} in 2022; the next solution appeared in 2023, it was hitchhiker's problem \#2, Klement et al.\ \cite{Klemetal}; and then, a possible solution for the bivariate case of hitchhiker's problem \#4 was given by Stopar \cite{St} in 2024. The hitchhiker's problem \#5 appears to be more involved. It asks to find the maximal negative and positive mass {over all boxes over all quasi-copulas} as well as to characterize the type of boxes that have the maximal mass. We remark that the term \textit{maximal negative} refers to the negative value with the largest absolute value. 
The problem started originally by Nelsen et al.\ in 2002 \cite{Nelsen2002}, where an analytical solution for the bivariate case is given. In 2007 de Baets et al.\  \cite{BMUF07} propose a linear programming  approach to give the trivariate case. In 2023 \' Ubeda-Flores \cite{UF23} extends the linear programming approach up to dimension four and conjectures the growth for higher dimensions. In our very recent work \cite{BeOmVuZa} we extended this approach further, up to dimension $17$ and disproved conjecture of \cite{UF23}. In this paper we give the final solution of the problem for all dimensions.
It turns out that there is no closed formula that applies simultaneously in all dimensions. However, there is an explicit formula for each dimension that requires performing a simple algorithm to determine the smallest positive integer that satisfies a particular inequality. This smallest integer then determines the maximal positive (resp.\ maximal negative) volume {over all boxes} over all quasi-copulas and also a realization of a box and a quasi-copula. The overall growth of the maximal volumes is exponential.
For the maximal positive value, the boxes $\cB$ are always of the form $[a,1]^d$, while for maximal negative values this applies to $d\geq 7$.

{
The paper is organized as follows. In Section \ref{sec:main-theorems}
we state our main results, that solve \cite[Open Problem 5]{ARIASGARCIA20201}
(see Theorems \ref{sol:min-value} and \ref{sol:max-value}).
Since there are no closed form solutions, we present a detailed numerical analysis of the solutions in Section \ref{sec:numerical-aspects}.
In Section \ref{preparatory} we formulate the maximal volume problems in terms of linear programs (see Proposition \ref{prel:prop}),
simplify them (see Proposition \ref{simplify-further}) and derive the dual linear programs (see Proposition \ref{prop:simplification}).
In Section \ref{sec:basic-technical} we establish the basic technical result (see Proposition \ref{auxiliary-proposition}),
which is then used in Sections \ref{sec:minimal} and \ref{sec:maximal}
 to solve the dual linear programs and thus prove Theorems
\ref{sol:min-value} and \ref{sol:max-value}.
Finally, in Section \ref{sec:remaining-cases},
we solve the remaining cases of the maximal negative volume problem in the dimensions $d=3,4,5,6$ analytically.
}


\section{Statements of the main theorems}
\label{sec:main-theorems}

In this section we introduce the notation and state our main results that
solve \cite[Open Problem 5]{ARIASGARCIA20201}, i.e., Theorem \ref{sol:min-value} solves the maximal negative volume problem, while Theorem \ref{sol:max-value} solves the maximal positive volume problem. Concrete examples are also presented (see Examples \ref{min-dim-7} and \ref{max-dim-8}).\\

Let $\calD\subseteq [0,1]^d$ be a set. We say that a function $Q:\calD\to [0,1]$ satisfies:
\begin{enumerate}
    \item 
    \label{BC}
    \emph{Boundary condition:}
        For any index $i=1, \ldots, d$ the following holds:
        \begin{align*}
        \text{(a)}\quad &\text{If }(u_1, \ldots, u_{i-1}, 0, u_{i+1}, \ldots, u_d)\in \calD,\text{ then } Q(u_1, \ldots, u_{i-1}, 0, u_{i+1}, \ldots, u_d) = 0.\\
        \text{(b)}\quad &\text{If }(1, \ldots, 1, u_i, 1, \ldots, 1)\in \calD,\text{ then } Q(1, \ldots, 1, u_i, 1, \ldots, 1) = u_i.
        \end{align*}
    \item
    \label{Mon}
    \emph{Monotonicity condition:} 
    $Q$ is nondecreasing in every variable, i.e., for each $i=1,\ldots,d$ and each pair of $d$-tuples
    \begin{align*}
    \begin{split}
    \uu&:=(u_1, \ldots,u_{i-1},u_i,u_{i+1},\ldots,u_d)\in \calD,\\
    \widetilde\uu&:=(u_1, \ldots,u_{i-1},\widetilde u_i,u_{i+1},\ldots,u_d)\in \calD,
    \end{split}
    \end{align*}
    such that $u_i\leq \widetilde u_i$, it follows that
    $Q(\uu)\leq Q(\widetilde\uu)$.
    \item 
    \label{LC}
    \emph{Lipschitz condition}:
        Given $d$--tuples 
        $(u_1, \ldots, u_d)$ and $(v_1, \ldots, v_d)$ 
        in $\calD$ it holds that
        \begin{align*}
        |Q(u_1, \ldots, u_d) - Q(v_1, \ldots, v_d)| \leq \sum_{i = 1}^{d}|u_i - v_i|.
        \end{align*}
\end{enumerate}
If $\calD=[0,1]^d$ and $Q$ satisfies \eqref{BC}, \eqref{Mon}, \eqref{LC}, then $Q$ is called a \emph{$d$-variate quasi-copula} (or \emph{$d$-quasi-copula}). We will omit the dimension $d$ when it is clear from the context and write quasi-copula for short.
\\

Let $Q$ be a quasi-copula and $\cB=\prod_{i=1}^d [a_i,b_i]\subseteq [0,1]$ a $d$-box with $a_i<b_i$
for each $i$.
We will use multi-indices
of the form $\mi := (\mi_1, \mi_2, \ldots, \mi_d) \in \{0, 1\}^d$ to index 
$2^d$ vertices $\prod_{i=1}^d\{a_i,b_i\}$
of $\cB$. 
We write 
\begin{equation*}
    x_\mi:=((x_\mi)_1,\ldots,(x_\mi)_d),\quad 
    \text{where}\quad
    (x_\mi)_k = \begin{cases}
    a_k,\quad \text{if }\mi_k=0,\\
    b_k,\quad \text{if }\mi_k=1.
    \end{cases}
\end{equation*}
Let us denote the value of $Q$ at the point $x_\mi$
by 
\begin{equation*}
    q_\mi := Q(x_\mi).
\end{equation*}
Let $\|\mi\|_1:=\sum_{i=1}^d \mi_i$ be the $1$-norm of the multi-index $\mi$
and $\sign(\mi) := (-1)^{d-\|\mi\|_1}$ its sign.
The \textbf{$Q$-volume} of $\cB$ is defined by:
\begin{equation}
    V_Q(\cB) = \sum_{\mi\in \{0, 1\}^d} \sign(\mi) q_\mi.
    \label{eq:volume}
\end{equation}

\bigskip

The complete solution to \cite[Open Problem 5]{ARIASGARCIA20201} for the maximal negative volume, i.e., the largest in absolute value among negative ones, is the following.

\begin{theorem}
\label{sol:min-value}
    Assume the notation above.
    Let $d\in \NN$ and $d\geq 7$.
Define
\begin{align*}
&(c_1,c_2,\ldots,c_{\lfloor \frac{d}{2}\rfloor}):=\\
=&
\left\{
    \begin{array}{rl}
    \Big(
        \binom{d-1}{0},
        \binom{d-1}{1},
        \binom{d-1}{2},
        \ldots,
        \binom{d-1}{\frac{d}{2}-1}\Big),&
            \text{if }d\text{ is even,}\\[1em]
    {\Big(
        \binom{d-1}{1},
        \binom{d-1}{1},
        \binom{d-1}{3},
        \binom{d-1}{3},
        \ldots,
        \binom{d-1}{\lfloor\frac{d}{2}\rfloor-1},
        \binom{d-1}{\lfloor\frac{d}{2}\rfloor-1}\Big)},&
            {\text{if  }d\text{ mod } 4=1,}\\[1em]    
    {\Big(
        \binom{d-1}{1},
        \binom{d-1}{1},
        \binom{d-1}{3},
        \binom{d-1}{3},
        \ldots,
        \binom{d-1}{\lfloor\frac{d}{2}\rfloor-2},
        \binom{d-1}{\lfloor\frac{d}{2}\rfloor-2},
        \binom{d-1}{\lfloor\frac{d}{2}\rfloor}\Big)},&
            {\text{if  }d\text{ mod } 4=3.}\\[1em]    
    \end{array}
\right.
\end{align*}
Define $c_0=0$
and
\begin{align*}
    w_{i,-}
    &:=\frac{1}{i+1}
    \Big(\sum_{j=0}^{i-1} c_{\lfloor\frac{d}{2}\rfloor-j} -1\Big) 
    \quad \text{for }i=1,\ldots,\lfloor\frac{d}{2}\rfloor.
\end{align*}
Let $i_0$ be the smallest integer in $\{1,\ldots,\lfloor\frac{d}{2}\rfloor\}$ such that
\begin{equation} \label{thm1:cond-to-check}
w_{i_0,-}\geq c_{\lfloor\frac{d}{2}\rfloor-i_0}.
\end{equation}
Then:
\begin{enumerate}
    \item The maximal negative volume $V_{Q}(\cB)$ of some box $\cB$       over all $d$-quasi-copulas $Q$ is equal to $-w_{i_0,-}$.
    \smallskip
    \item One of the realizations of $\cB$ and $Q$ is 
        $$
        \cB=\left[\frac{i_0}{i_0+1},1\right]^d,\quad
        q_{(0,\ldots,0)}=0,
        \quad
        q_\mi=\sum_{i=1}^{\|\mi\|_1} \delta_i\quad 
        \text{for all }\mi\in\{0,1\}^d\setminus \{0\}^d,
        $$
    where:
    \medskip
    \begin{enumerate}
    \item If $d$ is even, then for $j=1,\ldots,d$ we have
    \begin{align}
    \label{delta-min-optimal-even}
    \delta_{j}
    &=
    \left\{
        \begin{array}{rl}
            \frac{1}{i_0+1},& 
                \text{if }j=d\text{ or }(j\text{ is odd and }\binom{d-1}{j-1}>c_{\frac{d}{2}-i_0}),\\[0.5em]
            0,& \text{otherwise}.
        \end{array}
        \right.
    \end{align}
    \item If $d$ is odd, then for $j=1,\ldots,d$ we have
    \begin{align}
    \label{delta-min-optimal-odd}
    \delta_{j}
    &=
    \left\{
        \begin{array}{rl}
            \frac{1}{i_0+1},& 
                \text{if }j=d\text{ or }(j\text{ is even and }\binom{d-1}{j-1}>c_{\lfloor\frac{d}{2}\rfloor-i_0}),\\[0.5em]
            0,& \text{otherwise}.
        \end{array}
        \right.
    \end{align}
    \end{enumerate}
    This realization of $Q$ on 
    $\left\{\frac{i_0}{i_0+1},1\right\}^d$ indeed extends to a 
    quasi-copula $Q:[0,1]^d\to [0,1]$
    by \cite[Theorem 2.1]{BeOmVuZa}.
\end{enumerate}
\end{theorem}

\bigskip
Let us demonstrate the statement of Theorem \ref{sol:min-value} in the smallest dimension $d=7$.

\begin{example}
\label{min-dim-7}
Assume the notation from Theorem \ref{sol:min-value}.
Let $d=7$. Define
$$(c_1,c_2,c_3):=
    \Big(
        \binom{6}{1},
        \binom{6}{1},
        \binom{6}{3}
    \Big)=(6,6,20)$$
and $c_0=0$.
Let
\begin{align*}
    w_{1,-}&:=\frac{1}{2}(c_3-1)=\frac{19}{2},\\
    w_{2,-}&:=\frac{1}{3}(c_3+c_2-1)=\frac{25}{3},\\
    w_{3,-}&:=\frac{1}{4}(c_3+c_2+c_1-1)=\frac{31}{4}.
\end{align*}
Note that the smallest $i\in \{1,2,3\}$ such that $w_{i,-}\geq c_{3-i}$, is $i_0=1$, i.e.,
$w_{1,-}\geq c_2$.
By Theorem \ref{sol:min-value}, the maximal negative volume of some box $\cB$       over all $7$-quasi-copulas $Q$ is equal to 
    $$-w_{1,-}=-\frac{19}{2}.$$
Further on, 
\begin{align*}
    \delta_{j}
    &=
    \left\{
        \begin{array}{rl}
            \frac{1}{2},& 
                \text{if }j\in\{4,7\},\\[0.2em]
            0,& \text{if }j\in\{1,2,3,5,6\}.
        \end{array}
        \right.
\end{align*}
Hence, one of the realizations of $\cB$ and $Q$ is 
        $$
        \cB=\left[\frac{1}{2},1\right]^7,\quad
        q_{(0,\ldots,0)}=0,
        \quad
        q_\mi=
        \left\{
            \begin{array}{rl}
            0,& \text{if }\|\mi\|_1\in \{0,1,2,3\},\\[0.3em]
            \frac{1}{2},& \text{if }\|\mi\|_1\in \{4,5,6\},\\[0.3em]
            1,& \text{if }\mi=(1,\ldots,1).
            \end{array}
        \right.
        $$
Note that this realization agrees with the one obtained by computer software approach in \cite[Table 1]{BeOmVuZa}.
\end{example}


The complete solution to \cite[Open Problem 5]{ARIASGARCIA20201} for the maximal positive volume is the following.

\begin{theorem}
\label{sol:max-value}
   Assume the notation above.
    Let $d\in \NN$ and $d\geq 2$.
Define
\begin{align*}
&(c_1,c_2,\ldots,c_{\lfloor \frac{d+1}{2}\rfloor-1}):=\\
=&
\left\{
    \begin{array}{rl}
    \Big(
        \binom{d-1}{1},
        \binom{d-1}{2},
        \binom{d-1}{3},
        \ldots,
        \binom{d-1}{\frac{d}{2}-1}\Big),&
            \text{if }d\text{ is even,}\\[1em]
    {\Big(
        \binom{d-1}{0},
        \binom{d-1}{2},
        \binom{d-1}{2},
        \binom{d-1}{4},
        \binom{d-1}{4},
        \ldots,
        \binom{d-1}{\lfloor\frac{d}{2}\rfloor-2},
        \binom{d-1}{\lfloor\frac{d}{2}\rfloor-2},
        \binom{d-1}{\lfloor\frac{d}{2}\rfloor}\Big),}&
            {\text{if  }d\text{ mod } 4=1,}\\[1em]    
    {\Big(
        \binom{d-1}{0},
        \binom{d-1}{2},
        \binom{d-1}{2},
        \binom{d-1}{4},
        \binom{d-1}{4},
        \ldots,
        \binom{d-1}{\lfloor\frac{d}{2}\rfloor-1},
        \binom{d-1}{\lfloor\frac{d}{2}\rfloor-1}\Big),}&
            {\text{if  }d\text{ mod } 4=3.}\\[1em]    
    \end{array}
\right.
\end{align*}
Define $c_0=0$
and
\begin{align*}
    w_{i,+}
    &:=\frac{1}{i+1}
    \Big(\sum_{j=0}^{i-1} c_{\lfloor\frac{d+1}{2}\rfloor-1-j} +1\Big) 
    \quad \text{for }i=1,\ldots,\lfloor\frac{d+1}{2}\rfloor-1.
\end{align*}
Let $i_0$ be the smallest integer in $\{1,\ldots,\lfloor\frac{d+1}{2}\rfloor-1\}$ such that
\begin{equation} \label{thm2:cond-to-check}
w_{i_0,+}\geq c_{\lfloor\frac{d+1}{2}\rfloor-1-i_0}.
\end{equation}
Then:
\begin{enumerate}
    \item The maximal positive volume $V_{Q}(\cB)$ of some box $\cB$       over all $d$-quasi-copulas $Q$ is equal to $w_{i_0,+}$.
    \item One of the realizations of $\cB$ and $Q$ is 
        $$
        \cB=\left[\frac{i_0}{i_0+1},1\right]^d,\quad
        q_{(0,\ldots,0)}=0,
        \quad
        q_\mi=\sum_{i=1}^{\|\mi\|_1} \delta_i\quad 
        \text{for all }\mi\in\{0,1\}^d\setminus \{0\}^d,
        $$
    where:
    \medskip
    \begin{enumerate}
    \item If $d$ is even, then for $j=1,\ldots,d$ we have
    \begin{align}
    \label{delta-min-optimal-even-max}
    \delta_{j}
    &=
    \left\{
        \begin{array}{rl}
            \frac{1}{i_0+1},& 
                \text{if }j=d\text{ or }(j\text{ is even and }\binom{d-1}{j-1}>c_{\frac{d}{2}-1-i_0}),\\[0.5em]
            0,& \text{otherwise}.
        \end{array}
        \right.
    \end{align}
    \item If $d$ is odd, then for $j=1,\ldots,d$ we have
    \begin{align}
    \label{delta-min-optimal-odd-max}
    \delta_{j}
    &=
    \left\{
        \begin{array}{rl}
            \frac{1}{i_0+1},& 
                \text{if }j=d\text{ or }(j\text{ is odd and }\binom{d-1}{j-1}>c_{\lfloor\frac{d+1}{2}\rfloor-1-i_0}),\\[0.5em]
            0,& \text{otherwise}.
        \end{array}
        \right.
    \end{align}
    \end{enumerate}
    This realization of $Q$ on 
    $\left\{\frac{i_0}{i_0+1},1\right\}^d$ indeed extends to a 
    quasi-copula $Q:[0,1]^d\to [0,1]$
    by \cite[Theorem 2.1]{BeOmVuZa}.
\end{enumerate}
\end{theorem}

\bigskip

Let us demonstrate the statement of Theorem \ref{sol:min-value} in dimension $d=8$.

\begin{example}
\label{max-dim-8}
Assume the notation from Theorem \ref{sol:max-value}.
Let $d=8$. Define
$$(c_1,c_2,c_3):=
    \Big(
        \binom{7}{1},
        \binom{7}{2},
        \binom{7}{3}
    \Big)=(7,21,35)$$
and $c_0=0$.
Let
\begin{align*}
    w_{1,+}&:=\frac{1}{2}(c_3+1)=18,\\
    w_{2,+}&:=\frac{1}{3}(c_3+c_2+1)=19,\\
    w_{3,+}&:=\frac{1}{4}(c_3+c_2+c_1+1)=16.
\end{align*}
Note that the smallest $i\in \{1,2,3\}$ such that $w_{i,+}\geq c_{3-i}$ is $i_0=2$, i.e.,
$w_{2,+}\geq c_1$.
By Theorem \ref{sol:max-value}, the maximal positive volume of some box $\cB$       over all $8$-quasi-copulas $Q$ is equal to 
    $$w_{2,+}=19.$$
Further on, 
\begin{align*}
    \delta_{j}
    &=
    \left\{
        \begin{array}{rl}
            \frac{1}{3},& 
                \text{if }j\in\{4,6,8\},\\[0.2em]
            0,& \text{if }j\in\{1,2,3,5,7\}.
        \end{array}
        \right.
\end{align*}
Hence, one of the realizations of $\cB$ and $Q$ is 
        $$
        \cB=\left[\frac{2}{3},1\right]^8,\quad
        q_{(0,\ldots,0)}=0,
        \quad
        q_\mi=
        \left\{
            \begin{array}{rl}
            0,& \text{if }\|\mi\|_1\in \{0,1,2,3\},\\[0.3em]
            \frac{1}{3},& \text{if }\|\mi\|_1\in \{4,5\},\\[0.3em]
            \frac{2}{3},& \text{if }\|\mi\|_1\in \{6,7\},\\[0.3em]
            1,& \text{if }\mi=(1,\ldots,1).
            \end{array}
        \right.
        $$
Note that this realization agrees with the one obtained by computer software approach in \cite[Table 2]{BeOmVuZa}.
\end{example}

{
In the next section we will give a detailed numerical aspects of Theorems \ref{sol:min-value} and \ref{sol:max-value}.
In the following remark we comment on the general connection between them.

\begin{remark}
\label{remark-after-theorems}
\begin{enumerate}
\item
Let us first assume that $d$ is even. Then the vectors of the coefficients $c_i$ in Theorems 
\ref{sol:min-value} and \ref{sol:max-value}
differ only in the fact that in the minimal volume case there is also a binomial symbol $\binom{d-1}{0}$ at the beginning. This difference arises because when dualising the primal linear program, which is the formulation of the problem we use, the important constraints that have a true influence on the solution in both cases are those with positive lower bounds. In our case, the bounds are signed binomial coefficients (i.e., coefficients appearing at variables in the primal objective function). In the case of minimal volume, the pattern of signed coefficients starts with $+$ and alternates between signs, while in the case of maximal volume it starts with $-$ and alternates.
In the question for the maximal volume, the coefficient $\binom{d-1}{d-1}=\binom{d-1}{0}$ is missing,
as we use its constraint to get rid of one of the variables. We prove that we must have equality in the optimal solution in this constraint.

Let us now assume that $d$ is odd. Then the coefficient vectors $c_i$ in both theorems differ only in that in the minimal volume case all odd binomial symbols $\binom{d-1}{2i+1}$ occur, while in the maximal volume case all even binomial symbols $\binom{d-1}{2i}$ occur (with the exception of $\binom{d-1}{d-1}$, since we again use its constraint to get rid of one of the variables). This difference happens for exactly the same reason as for the even case explained in the first paragraph above, except that this time the sequences of the binomial symbols are also different. This is due to $d$ being odd and therefore
$\binom{d-1}{2i+1}=\binom{d-1}{d-2i}$, where $d-2i$ is odd again.
\smallskip
\item
\label{remark-after-theorems-pt2}
Let us comment on the dimension constraint $d\geq 7$, which appears in Theorem \ref{sol:min-value}, while it does not appear in Theorem \ref{sol:max-value}.
The reason for this difference is easiest to see in the feasibility regions of the dual linear programs (see \eqref{dual-LP-symmetric-v3} and \eqref{dual-LP-symmetric-v3-max}). In the case of the minimal volume, one of the constraints is $\ell_d-y_3\geq -1$, while in the case of the maximal volume it is $\ell_d-y_3\geq 1$. As all variables are nonnegative, in the case of the maximal volume the constraint 
$\ell_d\geq 1+y_3$ is more restrictive than $\ell_d\geq 0$, while in the case of the minimal volume the constraint $\ell_d\geq 0$ could be more restrictive than the constraint $\ell_d\geq -1+y_3$. When analysing both possibilities, it turns out that we actually obtain an optimal solution for $d\leq 6$ if we are on the hyperplane $\ell_d=0$,
 in contrast to the cases $d\geq 7$, where none of the points with $\ell_d=0$
 lie in a feasibility region.
 \item 
Let us comment on the realizations of the boxes $\cB$ and the quasi-copula $Q$, given by
 Theorem \ref{sol:min-value} and \ref{sol:max-value}.
 As already observed in \cite[Remark 3.3]{BeOmVuZa}, the realizations of the solution are not unique.
 In the case of a non-symmetric solution to \eqref{LP} (i.e., $\|\mi\|_1=\|\mj\|_1$ does not imply $q_{\mi}=q_{\mj}$ or $a_i$ are not all equal or $b_i$ are not all equal), 
 many new solutions can be obtained by the action
 of the symmetric group
 (see the proof of Proposition \ref{symmetric-solution}).
 We can therefore form an equivalence relation $\sim$ on the set of all solutions.
 Two solutions are in relation if one can obtain the other by the action of an element of the symmetric group on the first (see \eqref{one-nonsymmetric-solution} and \eqref{second-nonsymmetric-solution}).
 If we take the average (see \eqref{average}) of all solutions in each equivalence class, we obtain the unique symmetric solution given in Theorems \ref{sol:min-value} and \ref{sol:max-value}. In this sense, this unique symmetric solution is the canonical solution.
 \end{enumerate}
\end{remark}


\section{Numerical aspects of Theorems \ref{sol:min-value}
and \ref{sol:max-value}}
\label{sec:numerical-aspects}

{
In this section we present various numerical aspects of Theorems \ref{sol:min-value}
and \ref{sol:max-value}\footnote{The numerical analysis in arithmetic over $\mathbb{Q}$ was performed using the software tool \textit{Mathematica} \cite{ram2024}. The source code is available at  \url{https://github.com/ZalarA/Quasi-copulas-extreme-volumes}.}. In Tables \ref{table-1} and \ref{table-2}  
we give maximal positive and negative volumes of the $d$-boxes over all $d$-quasi-copuas for dimensions up to 68.
(This is not due to the limitations of the computer software but for paper space reasons.) We also state the numbers $i_0$ were the procedure in Theorems \ref{sol:min-value}, \ref{sol:max-value} terminates.
Figures \ref{fig:figure1}, \ref{fig:figure2}, \ref{fig:figure3} demonstrate the growth of the number $i_0$ for the positive and the negative volume case, as well as their difference, respectively.
Figure \ref{fig:figure4} graphically represents the volumes of the solutions for both cases in the logarithmic scale with a base 2. We comment on the behaviour of the volumes in Remark \ref{rem:positive-negative-volumes} and on the behaviour of the difference in the values of $i_0$ in Remark \ref{rem:difference-i0}. Proposition \ref{dependence-max-min} states two possible regimes of the difference in even dimensions. 
}

\newpage

\begin{table}[h!]
      \caption{
      {
      Minimal values of $V_Q(\cB)$ over all $d$--variate quasi--copulas $Q$ and all $d$--boxes} $\cB\subseteq [0,1]^d$.
      {It turns out that for $d\geq 7$ 
      the minimal box $B_{\min}$ is of the form $[\frac{i_0}{i_0+1},1]^d$.}
      Dimensions $d<7$ are treated separately in Section \ref{sec:remaining-cases}.
      }
{
{
\renewcommand{\arraystretch}{1.5}
\begin{tabular}[t]{|c|c|c|}
\hline
 $d$ & $i_0$ & $V_Q\big([\frac{i_0}{i_0+1},1]^d\big)$ \\[0.2em]\hline\hline
 7 & 1 & $-{19\big/2}$ \\[0.2em]\hline
 8 & 2 & $-{55\big/3}$ \\[0.2em]\hline
 9 & 2 & $-37$ \\[0.2em]\hline
 10 & 2 & $-{209\big/3}$ \\[0.2em]\hline
 11 & 1 & $-{251\big/2}$ \\[0.2em]\hline
 12 & 2 & $-{791\big/3}$ \\[0.2em]\hline
 13 & 2 & $-{1583\big/3}$ \\[0.2em]\hline
 14 & 2 & $-{3002\big/3}$ \\[0.2em]\hline
 15 & 3 & $-{7435\big/4}$ \\[0.2em]\hline
 16 & 2 & $-3813$ \\[0.2em]\hline
 17 & 2 & $-{22879\big/3}$ \\[0.2em]\hline
 18 & 2 & $-{43757\big/3}$ \\[0.2em]\hline
 19 & 3 & $-{112267\big/4}$ \\[0.2em]\hline
 20 & 2 & $-{167959\big/3}$ \\[0.2em]\hline
 21 & 2 & $-111973$ \\[0.2em]\hline
 22 & 2 & $-{646645\big/3}$ \\[0.2em]\hline
 23 & 3 & $-{1700271\big/4}$ \\[0.2em]\hline
 24 & 2 & $-{2496143\big/3}$ \\[0.2em]\hline
 25 & 2 & $-{4992287\big/3}$ \\[0.2em]\hline
 26 & 3 & $-{12926459\big/4}$ \\[0.2em]\hline
 27 & 3 & $-{25852919\big/4}$ \\[0.2em]\hline
 \end{tabular}
 }
 \hfill
 }
 {
 {
\renewcommand{\arraystretch}{1.5}
 \begin{tabular}[t]{|c|c|c|}
\hline
 $d$ & $i_0$ & $V_Q\big([\frac{i_0}{i_0+1},1]^d\big)$  \\[0.2em]\hline\hline
 28 & 3 & $-{25240027\big/2}$ \\[0.2em]\hline
 29 & 2 & $-{74884319\big/3}$ \\[0.2em]\hline
 30 & 3 & $-{197318609\big/4}$ \\[0.2em]\hline
 31 & 3 & $-{394637219\big/4}$ \\[0.2em]\hline
 32 & 3 & $-{385987897\big/2}$ \\[0.2em]\hline
 33 & 2 & $-{1131445439\big/3}$ \\[0.2em]\hline
 34 & 3 & $-{3022770629\big/4}$ \\[0.2em]\hline
 35 & 3 & $-{6045541259\big/4}$ \\[0.2em]\hline
 36 & 3 & $-{11845439759\big/4}$ \\[0.2em]\hline
 37 & 2 & $-{17194993199\big/3}$ \\[0.2em]\hline
 38 & 3 & $-{46453775279\big/4}$ \\[0.2em]\hline
 39 & 3 & $-{92907550559\big/4}$ \\[0.2em]\hline
 40 & 3 & $-{182303526209\big/4}$ \\[0.2em]\hline
 41 & 4 & $-{440029574399\big/5}$ \\[0.2em]\hline
 42 & 3 & $-{715904248019\big/4}$ \\[0.2em]\hline
 43 & 3 & $-{1431808496039\big/4}$ \\[0.2em]\hline
 44 & 3 & $-{2813088831929\big/4}$ \\[0.2em]\hline
 45 & 4 & $-{6842895760271\big/5}$ \\[0.2em]\hline
 46 & 3 & $-{11060241944075\big/4}$ \\[0.2em]\hline
 47 & 3 & $-{22120483888151\big/4}$ \\[0.2em]\hline
 48 & 3 & $-{43509459123197\big/4}$ \\[0.2em]\hline
\end{tabular}}
\hfill
{
 {
\renewcommand{\arraystretch}{1.5}
\begin{tabular}[t]{|c|c|c|}
\hline
    $d$ & $i_0$ & $V_Q\big([\frac{i_0}{i_0+1},1]^d\big)$ \\[0.2em]\hline\hline
 49 & 4 & $-{106543877604607\big/5}$ \\[0.2em]\hline
 50 & 3 & $-{171248556584475\big/4}$ \\[0.2em]\hline
 51 & 3 & $-{342497113168951\big/4}$ \\[0.2em]\hline
 52 & 3 & $-{674344248506151\big/4}$ \\[0.2em]\hline
 53 & 4 & $-{1660843019667103\big/5}$ \\[0.2em]\hline
 54 & 3 & $-{2656661186721807\big/4}$ \\[0.2em]\hline
 55 & 3 & $-{5313322373443615\big/4}$ \\[0.2em]\hline
 56 & 3 & $-{10470793684489291\big/4}$ \\[0.2em]\hline
 57 & 4 & $-{25918766458461183\big/5}$ \\[0.2em]\hline
 58 & 3 & $-{41285913247219255\big/4}$ \\[0.2em]\hline
 59 & 3 & $-{82571826494438511\big/4}$ \\[0.2em]\hline
 60 & 3 & $-{81426118154032613\big/2}$ \\[0.2em]\hline
 61 & 4 & $-{404908794653887719\big/5}$ \\[0.2em]\hline
 62 & 3 & $-{642608465005843433\big/4}$ \\[0.2em]\hline
 63 & 3 & $-{1285216930011686867\big/4}$ \\[0.2em]\hline
 64 & 3 & $-{1268300040500821345\big/2}$ \\[0.2em]\hline
 65 & 4 & $-1266363348322336051$ \\[0.2em]\hline
 66 & 4 & $-2504757439320041105$ \\[0.2em]\hline
 67 & 3 & $-{20032397217749335691/4}$ \\[0.2em]\hline
 68 & 4 & $-{49552967470135840983/5}$ \\[0.2em]\hline
\end{tabular}
}}

}
\label{table-1}
\end{table}


\newpage

\begin{table}[h!]
      \caption{
      {
      Maximal 
      values of $V_Q(\cB)$ over all $d$--variate quasi--copulas $Q$ and all $d$--boxes} $\cB\subseteq [0,1]^d$.
      {It turns out that for $d\geq 3$ 
      the maximal box $B_{\max}$ is of the form $[\frac{i_0}{i_0+1},1]^d$.}
      }
{
{
\renewcommand{\arraystretch}{1.5}
\begin{tabular}[t]{|c|c|c|}
\hline
 $d$ & $i_0$ & $V_Q\big([\frac{i_0}{i_0+1},1]^d\big)$ \\[0.2em]\hline\hline
 3 & 1 & $ 1 $ \\[0.2em]\hline
 4 & 1 & $2 $ \\[0.2em]\hline
 5 & 1 & ${7\big/2} $ \\[0.2em]\hline
 6 & 1 & ${11\big/2} $ \\[0.2em]\hline
 7 & 2 & ${31\big/3} $ \\[0.2em]\hline
 8 & 2 & $ 19 $ \\[0.2em]\hline
 9 & 1 & ${71\big/2} $ \\[0.2em]\hline
 10 & 2 & ${211\big/3} $ \\[0.2em]\hline
 11 & 2 & ${421\big/3} $ \\[0.2em]\hline
 12 & 2 & ${793\big/3} $ \\[0.2em]\hline
 13 & 3 & ${1915\big/4} $ \\[0.2em]\hline
 14 & 2 & ${3004\big/3} $ \\[0.2em]\hline
 15 & 2 & ${6007\big/3} $ \\[0.2em]\hline
 16 & 2 & ${11441\big/3} $ \\[0.2em]\hline
 17 & 3 & ${28887\big/4} $ \\[0.2em]\hline
 18 & 2 & ${43759\big/3} $ \\[0.2em]\hline
 19 & 2 & ${87517\big/3} $ \\[0.2em]\hline
 20 & 2 & $55987 $ \\[0.2em]\hline
 21 & 3 & ${436697\big/4} $ \\[0.2em]\hline
 22 & 2 & $ 215549 $ \\[0.2em]\hline
 23 & 2 & ${1293293\big/3} $ \\[0.2em]\hline
 24 & 2 & ${2496145\big/3} $ \\[0.2em]\hline
 \end{tabular}
 }
 \hfill
 }
 {
 {
\renewcommand{\arraystretch}{1.5}
 \begin{tabular}[t]{|c|c|c|}
\hline
 $d$ & $i_0$ & $V_Q\big([\frac{i_0}{i_0+1},1]^d\big)$  \\[0.2em]\hline\hline
 25 & 3 & ${6626669\big/4} $ \\[0.2em]\hline
 26 & 3 & ${12926461\big/4} $ \\[0.2em]\hline
 27 & 2 & $6438467 $ \\[0.2em]\hline
 28 & 3 & $12620014 $ \\[0.2em]\hline
 29 & 3 & ${100960111\big/4} $ \\[0.2em]\hline
 30 & 3 & ${197318611\big/4} $ \\[0.2em]\hline
 31 & 2 & ${290845351\big/3} $ \\[0.2em]\hline
 32 & 3 & $192993949 $ \\[0.2em]\hline
 33 & 3 & ${1543951591\big/4} $ \\[0.2em]\hline
 34 & 3 & ${3022770631\big/4} $ \\[0.2em]\hline
 35 & 2 & ${4407922861\big/3} $ \\[0.2em]\hline
 36 & 3 & ${11845439761\big/4} $ \\[0.2em]\hline
 37 & 3 & ${23690879521\big/4} $ \\[0.2em]\hline
 38 & 3 & ${46453775281\big/4} $ \\[0.2em]\hline
 39 & 2 & ${67156001221\big/3} $ \\[0.2em]\hline
 40 & 3 & ${182303526211\big/4} $ \\[0.2em]\hline
 41 & 3 & ${364607052421\big/4} $ \\[0.2em]\hline
 42 & 3 & ${715904248021\big/4} $ \\[0.2em]\hline
 43 & 4 & ${1734977456941\big/5} $ \\[0.2em]\hline
 44 & 3 & ${2813088831931\big/4} $ \\[0.2em]\hline
 45 & 3 & ${5626177663861\big/4} $ \\[0.2em]\hline
 46 & 3 & ${11060241944077\big/4} $ \\[0.2em]\hline
\end{tabular}}
\hfill
}
{
 {
\renewcommand{\arraystretch}{1.5}
\begin{tabular}[t]{|c|c|c|}
\hline
    $d$ & $i_0$ & $V_Q\big([\frac{i_0}{i_0+1},1]^d\big)$ \\[0.2em]\hline\hline
     47 & 4 & ${26997208242193\big/5} $ \\[0.2em]\hline
 48 & 3 & ${43509459123199\big/4} $ \\[0.2em]\hline
 49 & 3 & ${87018918246397\big/4} $ \\[0.2em]\hline
 50 & 3 & ${171248556584477\big/4} $ \\[0.2em]\hline
 51 & 4 & ${420596950601801\big/5} $ \\[0.2em]\hline
 52 & 3 & ${674344248506153\big/4} $ \\[0.2em]\hline
 53 & 3 & ${1348688497012305\big/4} $ \\[0.2em]\hline
 54 & 3 & ${2656661186721809\big/4} $ \\[0.2em]\hline
 55 & 4 & ${6560130872627137\big/5} $ \\[0.2em]\hline
 56 & 3 & ${10470793684489293\big/4} $ \\[0.2em]\hline
 57 & 3 & ${20941587368978585\big/4} $ \\[0.2em]\hline
 58 & 3 & ${41285913247219257\big/4} $ \\[0.2em]\hline
 59 & 4 & ${102430771061474087\big/5} $ \\[0.2em]\hline
 60 & 3 & $40713059077016307 $ \\[0.2em]\hline
 61 & 3 & ${325704472616130455\big/4} $ \\[0.2em]\hline
 62 & 3 & ${642608465005843435\big/4} $ \\[0.2em]\hline
 63 & 4 & ${1600998867547843289\big/5} $ \\[0.2em]\hline
 64 & 3 & $634150020250410673 $ \\[0.2em]\hline
 65 & 3 & ${5073200162003285383\big/4} $ \\[0.2em]\hline
 66 & 4 & ${12523787196600205527\big/5} $ \\[0.2em]\hline
 67 & 4 & ${25047574393200411053\big/5} $ \\[0.2em]\hline
 68 & 4 & $9910593494027168197 $ \\[0.2em]\hline
\end{tabular}}
}
\label{table-2}
\end{table}


\newpage

\begin{center}
\begin{figure}[h!]
      \caption{
      {
      Points represent the pairs $(d,i_0)$ such that a $d$-box $\cB=\big[\frac{i_0}{i_0+1},1\big]^d$, obtained by Theorem \ref{sol:min-value}, is a maximal negative volume $V_Q(\cB)$ box in  dimension $d$ over all $d$-quasi-copulas and over all $d$-boxes.}}
\includegraphics[width=14cm]{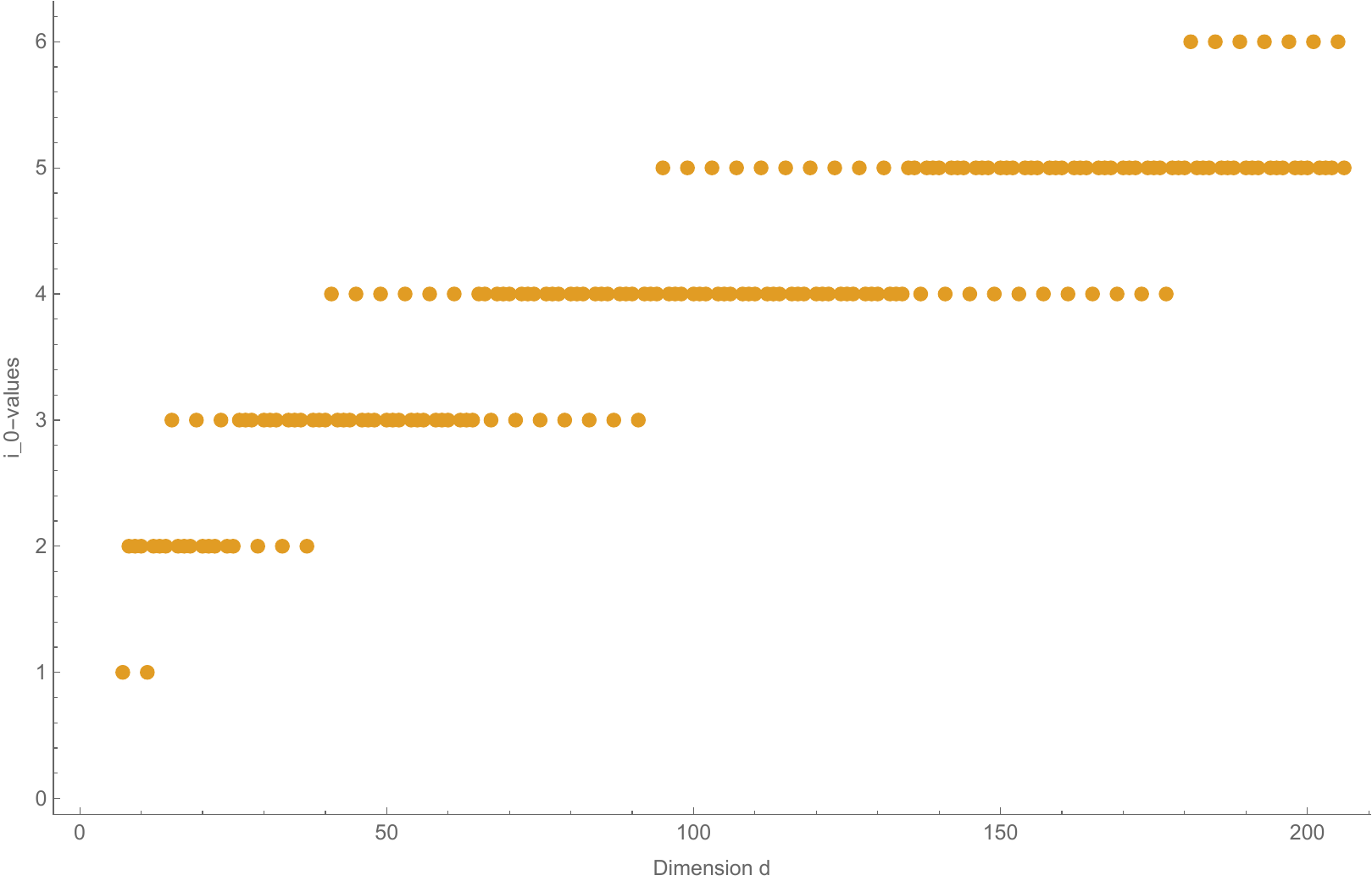}
\label{fig:figure1}
\end{figure}
\end{center}

\begin{center}
\begin{figure}[h!]
      \caption{
      {
      Points represent the pairs $(d,i_0)$ such that a $d$-box $\cB=\big[\frac{i_0}{i_0+1},1\big]^d$, obtained by Theorem \ref{sol:max-value}, is a maximal positive volume $V_Q(\cB)$ box in  dimension $d$ over all $d$-quasi-copulas and over all $d$-boxes.}}
\includegraphics[width=14.1cm]{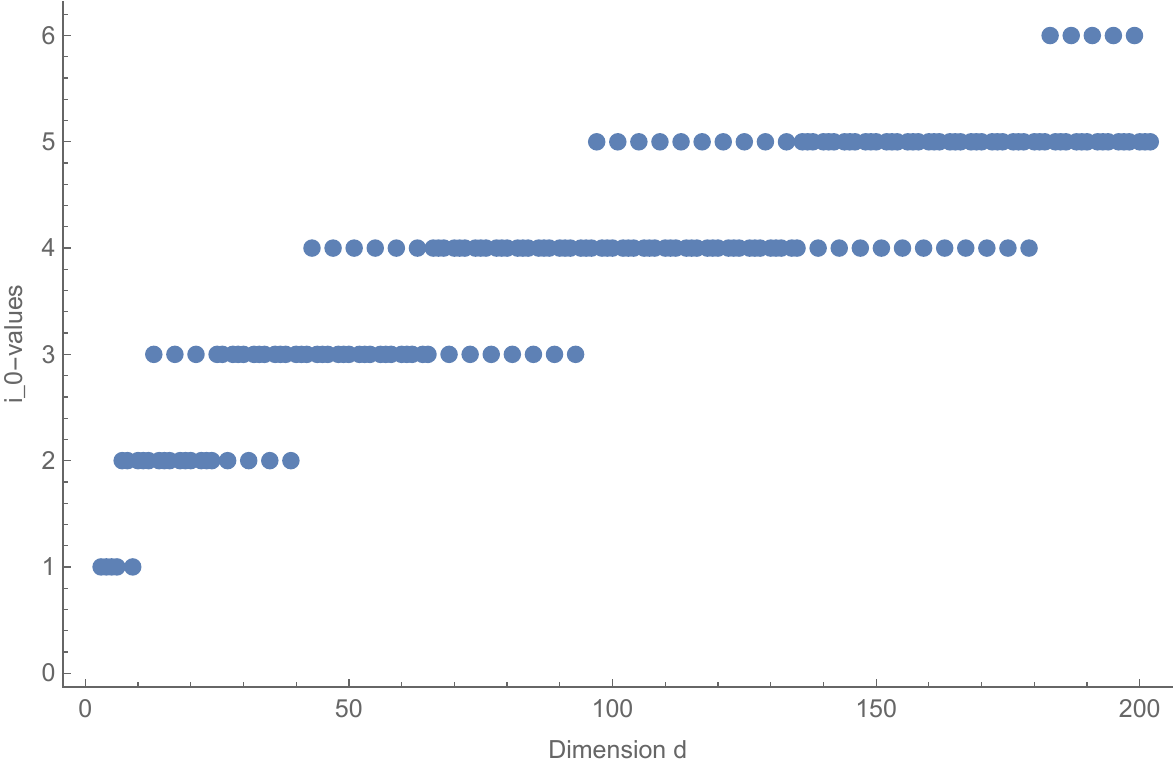}
\label{fig:figure2}
\end{figure}
\end{center}

\begin{center}
\begin{figure}[h!]
      \caption{
      {
      Points represent the pairs $(d,j)$, where $d$ is the dimension and $j$ is the difference $(i_0)_{\max}-(i_0)_{\min}$ such that a $d$-box $\cB_{\max}=\big[\frac{(i_0)_{\max}}{(i_0)_{\max}+1},1\big]^d$,
       obtained by Theorem \ref{sol:min-value},
      (resp.\ $\cB_{\min}=\big[\frac{(i_0)_{\min}}{(i_0)_{\min}+1},1\big]^d$,
      obtained by Theorem \ref{sol:max-value})
      is a maximal positive volume $V_Q(\cB_{\max})$
      (resp.\ a maximal negative volume $V_Q(\cB_{\min})$)
      box in  dimension $d$ over all $d$-quasi-copulas and over all $d$-boxes.}}
\includegraphics[width=14.1cm]{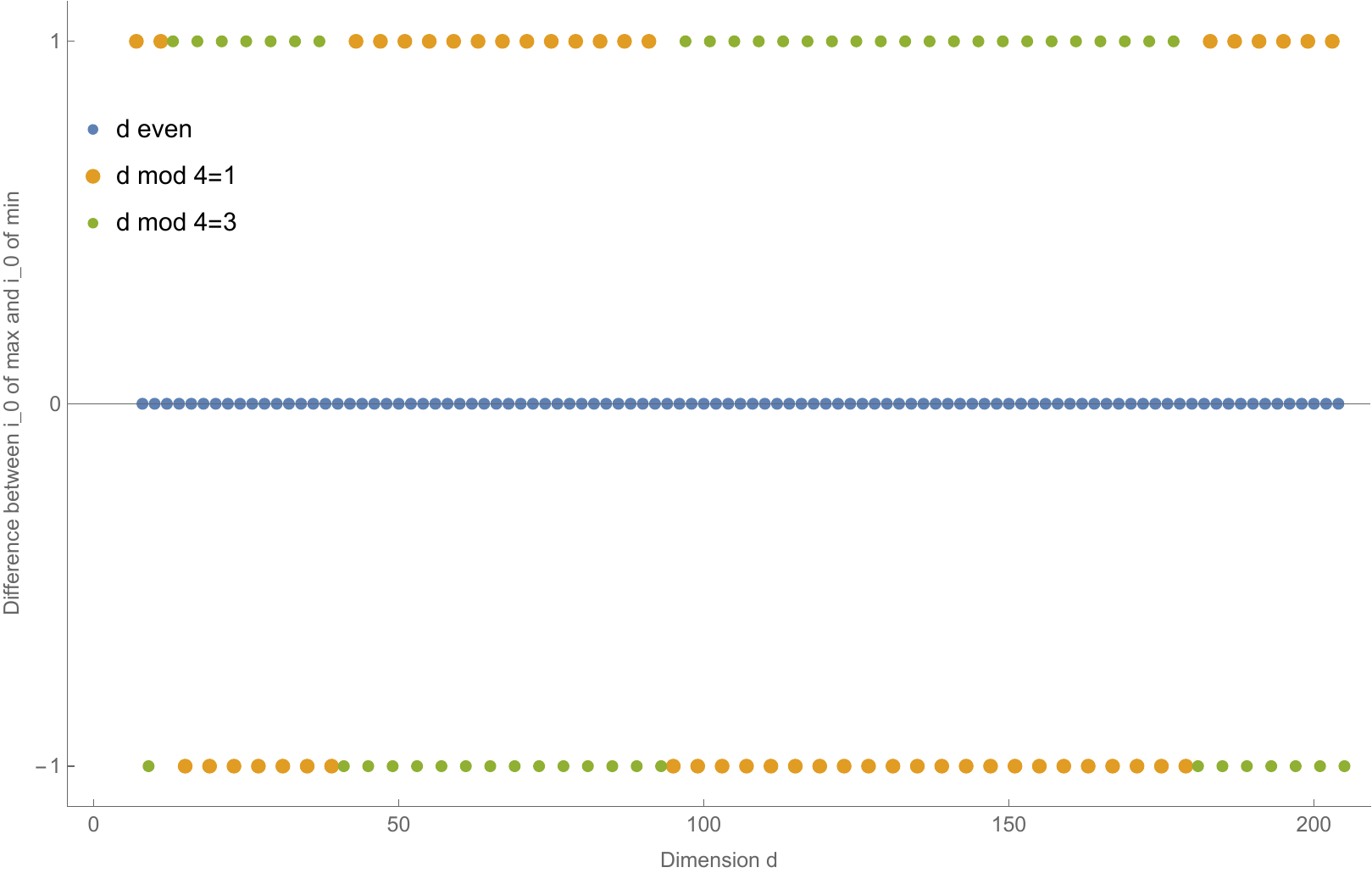}
\label{fig:figure3}
\end{figure}
\end{center}

\begin{center}
\begin{figure}[h!]
      \caption{
      {Blue (and orange) points represent the volumes of boxes with maximal (and minimal) volume over all $d$-quasi-copulas and over all $d$-boxes. The $x$-axis represents the dimension $d$, while the $y$-axis represents the value of the volume in the logarithmic scale with base 2.}}
\includegraphics[width=14.3cm]{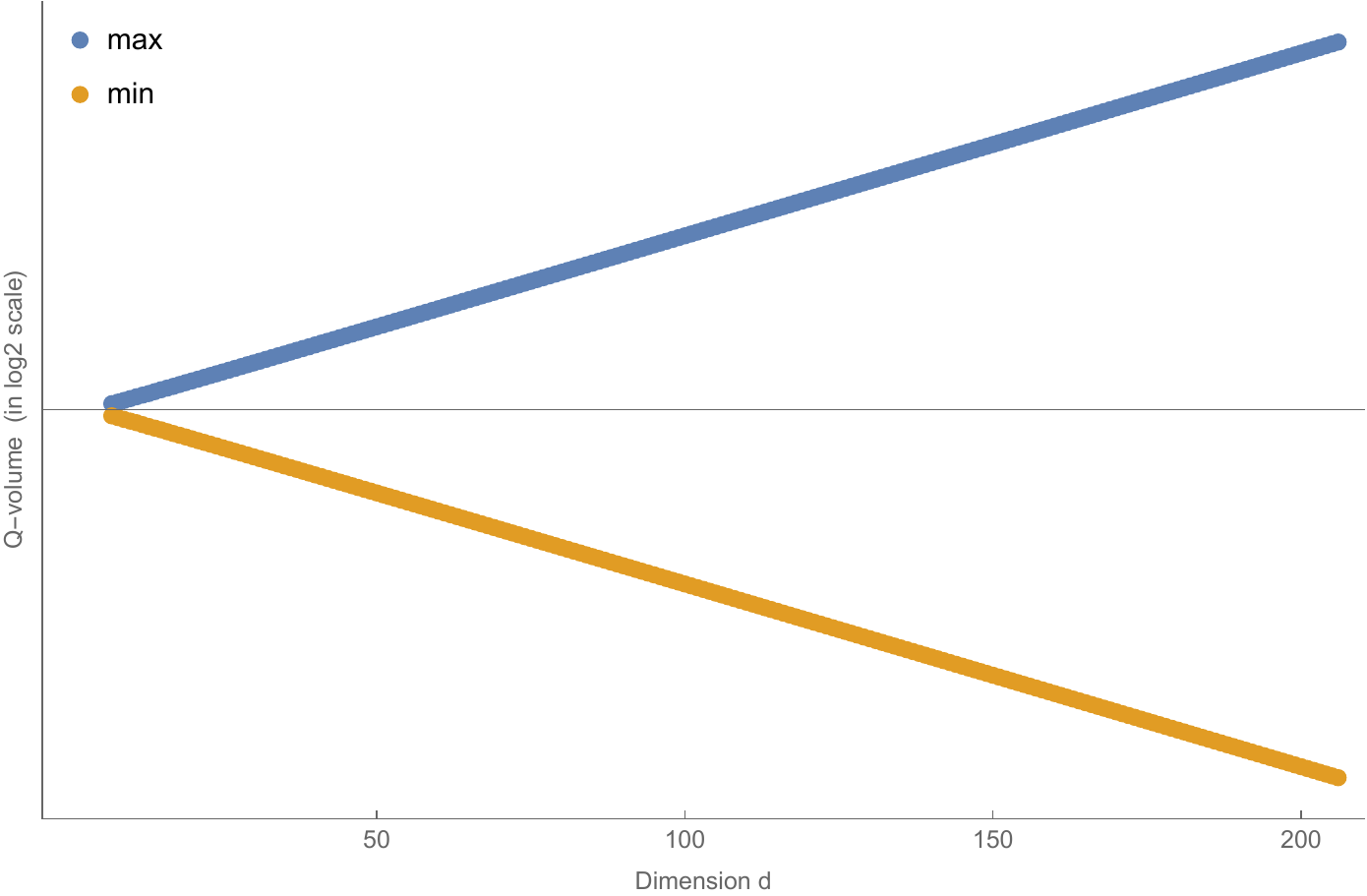}
\label{fig:figure4}
\end{figure}
\end{center}

{
\begin{remark}
\label{rem:positive-negative-volumes}
Let us comment on the values of the solutions $w_{i_0,+}$ and $w_{i_0,-}$
in Theorem \ref{sol:min-value} and \ref{sol:max-value}.

Assume that $d$ is even. Then
$$
    w_{\frac{d}{2},-}
    =
    \frac{1}{\frac{d}{2}+1}
        \Big[\sum_{j=0}^{\frac{d}{2}-1}\binom{d-1}{j}-1\Big]
    =
    \frac{1}{\frac{d}{2}+1}
        (2^{d-2}-1),
$$
since the sum of binomial symbols in brackets is precisely half of the sum 
$(1+1)^{d-1}=\sum_{j=0}^{d-1}\binom{d-1}{j}.$
Similarly, we have
$$
    w_{\frac{d}{2}-1,+}
    =
    \frac{1}{\frac{d}{2}}
        \Big[\sum_{j=1}^{\frac{d}{2}-1}\binom{d-1}{j}+1\Big]
    =
    \frac{1}{d}
        2^{d-1}.
$$

Assume now that $d$ is odd. In 
    $w_{\lfloor\frac{d}{2}\rfloor,-}$
the sum of all odd binomial coefficients $\binom{d-1}{2j+1}$ appears, which is equal to $2^{d-2}$, and hence 
$w_{\lfloor\frac{d}{2}\rfloor,-}=\frac{1}{\lfloor \frac{d}{2}\rfloor+1}(2^{d-2}-1)$.
In $w_{\lfloor\frac{d}{2}\rfloor,+}$
the sum of all even binomial coefficients $\binom{d-1}{2j}$ but $\binom{d-1}{d-1}$ appears, and hence 
$w_{\lfloor\frac{d}{2}\rfloor,+}=\frac{1}{\lfloor \frac{d+1}{2}\rfloor}2^{d-2}$.

So if the iterative procedures in Theorems \ref{sol:min-value} and \ref{sol:max-value} terminate after the maximal number of steps, the absolute volumes are as stated above. However, in practice the procedure terminates much earlier, since the sum of the binomial coefficients $\sum_{j=j_0}^{\frac{d}{2}-1}\binom{d-1}{j}$ is large in comparison to $\binom{d-1}{j_0-1}$ for relatively large $j_0$ and thus relatively small $i_0$. Observing Figures \ref{fig:figure1} and \ref{fig:figure2}
we see that $i_0$ is almost increasing function (with some exceptions) and its maximal value in dimensions up to $d=200$ is 6.
Experimentally, for a fixed $d$, we plot the function $w_{i,\pm}$,
for $i=1,\ldots,\lfloor\frac{d}{2}\rfloor$,
and the value $w_{\lfloor\frac{d}{2}\rfloor,\pm}$ was always the smallest.
However, proving this rigorously would require involved analysis of the behaviour of binomial symbols and their sums. Moreover, this would not give actual behaviour of the extreme volume solutions, since we do not know how $i_0$ in Theorems  \ref{sol:min-value} and \ref{sol:max-value} behaves. Based on Figure \ref{fig:figure4}, the actual volumes are of the form $2^{d-k}$ for some small $k$ relative to $d$ which is behaving similarly as $i$ (i.e., growing with the dimension).
\end{remark}
}

\bigskip

The next proposition states some relations among
    $i_0^{(\min)}$, $i_0^{(\max)}$, $w_{i_0^{(\min)},-}$ and $w_{i_0^{(\max)},+}$
from Theorems \ref{sol:min-value} and \ref{sol:max-value} in even dimensions.

{
\begin{proposition}\label{dependence-max-min}
Let $d\in 2\NN$ with $d\geq 8$. Assume the notation from Theorems \ref{sol:min-value} and \ref{sol:max-value}. Let $i_0^{(\min)}$ and $i_0^{(\max)}$
be the integers representing $i_0$ in Theorems \ref{sol:min-value} and \ref{sol:max-value}, respectively. Then the following statements hold:
\begin{enumerate}
\item\label{dependence-max-min-pt1} $0\leq i_0^{(\min)}-i_0^{(\max)}\leq 1$.
\smallskip
\item\label{dependence-max-min-pt2} If $i_0^{(\min)}=i_0^{(\max)}=:I$, then 
    $w_{i_0^{(\max)},+}=w_{i_0^{(\min)},-}+\frac{2}{I+1}$.
\end{enumerate}
\end{proposition}

\begin{proof}
Observe that by definition of $w_{i,+}, w_{i,-}$, we have that $w_{i,+}=w_{i,-}+\frac{2}{i+1}$.
In particular, this proves \eqref{dependence-max-min-pt2}.
Further, the lower bounds in the conditions \eqref{thm1:cond-to-check} and \eqref{thm2:cond-to-check} are the same, because $d$ is even.
Therefore if $i$ satisfies \eqref{thm1:cond-to-check}, it also satisfies \eqref{thm2:cond-to-check}. This proves the left inequality in \eqref{dependence-max-min-pt1}, i.e., $0\leq i_0^{(\min)}-i_0^{(\max)}$.
It remains to prove $i_0^{(\min)}-i_0^{(\max)}\leq 1.$ Assume that 
$i_0^{(\min)}>i_0^{(\max)}$. This means that 
\begin{equation}
\label{relations}
w_{i_0^{(\min)},-}<\binom{d-1}{\frac{d}{2}-1-i_0^{(\max)}}\leq w_{i_0^{(\max)},+}.
\end{equation}
We have to prove that $i_0^{(\min)}=i_0^{(\max)}+1$ or equivalently
\begin{equation}
\label{to-prove-v1}
\binom{d-1}{\frac{d}{2}-2-i_0^{(\max)}}\leq w_{i_0^{(\max)}+1,-}.
\end{equation}
Further, \eqref{to-prove-v1} is equivalent to
\begin{equation}
\label{to-prove-v2}
(i_0^{(\max)}+2)\binom{d-1}{\frac{d}{2}-2-i_0^{(\max)}}+1
\leq 
\sum_{j=0}^{i_0^{(\max)}}\binom{d-1}{\frac{d}{2}-1-j}.
\end{equation}
Note that the right inequality in \eqref{relations} is equivalent to 
\begin{equation}
\label{relations-v2}
    (i_0^{(\max)}+1)\binom{d-1}{\frac{d}{2}-1-i_0^{(\max)}}-1
    \leq 
    \sum_{j=0}^{i_0^{(\max)}-1}\binom{d-1}{\frac{d}{2}-1-j}.
\end{equation}
Using \eqref{relations-v2} to bound below the right hand side of \eqref{to-prove-v2},  \eqref{to-prove-v2} will follow if we prove that 
\begin{equation}
\label{to-prove-v3}
(i_0^{(\max)}+2)\binom{d-1}{\frac{d}{2}-2-i_0^{(\max)}}+1
\leq 
(i_0^{(\max)}+1)\binom{d-1}{\frac{d}{2}-1-i_0^{(\max)}}-1
+\binom{d-1}{\frac{d}{2}-1-i_0^{(\max)}}
\end{equation}
or equivalently
\begin{equation}
\label{to-prove-v4}
2
\leq 
(i_0^{(\max)}+2)
\left[
\binom{d-1}{\frac{d}{2}-1-i_0^{(\max)}}
-\binom{d-1}{\frac{d}{2}-2-i_0^{(\max)}}
\right]
\end{equation}
But this inequality clearly holds, since both brackets are natural numbers larger that $1$.
\end{proof}

\begin{remark}
\label{rem:difference-i0}
\begin{enumerate}[leftmargin=*]
\item 
Proposition \ref{dependence-max-min} states that the solutions to the maximal positive and maximal negative volume box question in even dimension, given by iterative procedures in Theorems \ref{sol:min-value} and \ref{sol:max-value},
are obtained after the same number of steps or after at most 1 more step for the maximal negative volume question. If both procedures end after the same number of steps $I$, then the maximal positive volume is larger for $\frac{2}{I+1}$ from the absolute value of the maximal negative volume. Moreover, the boxes realizing the solutions are then the same for both problems.
\item For odd $d$ the connection between the maximal volume box and the minimal volume box is more subtle. Observe that among all binomial symbols $\binom{d-1}{i}$, every symbol $\binom{d-1}{i}$ with odd $i$ appears
in the statement of Theorem \ref{sol:min-value}, while every symbol $\binom{d-1}{i}$
with even $i$ 
in the statement of Theorem \ref{sol:max-value}.
Hence, it is difficult to observe some clear connection between the solutions.
However, we observe that (see Figure \ref{fig:figure1}) the difference between 
$i_0^{(\min)}$ and $i_0^{(\max)}$, which represent $i_0$ in Theorems \ref{sol:min-value} and \ref{sol:max-value}, respectively, seems to always be either $1$ or $-1$.
The sign does not depend on $i \text{ mod }4$ but behaves somewhat randomly.
\end{enumerate}
\end{remark}



}


\section{Towards the proof of Theorems \ref{sol:min-value} and \ref{sol:max-value}}
\label{preparatory}

In this section we first recall the formulation of the maximal volume problems 
from \cite{BeOmVuZa} in terms of the linear programs (see Proposition \ref{prel:prop}). 
Then we simplify the linear programs by using symmetries in the variables (see Proposition \ref{symmetric-solution} and Corollary \ref{cor:sim}).
Next, the simplified linear programs are further reduced to smaller ones 
observing redundancy of some conditions (see Proposition \ref{simplify-further}).
Finally, the dual linear programs, which need to be solved, are obtained
(see Proposition \ref{prop:simplification}). We use them to prove that the proposed solution is indeed optimal.\\

\subsection{Notation and preliminaries}
Fix $d\in \NN$. 
For multi-indices 
$$
\mi=(\mi_1,\ldots,\mi_d)\in \{0,1\}^d\quad \text{and} 
\quad
\mj=(\mj_1,\ldots,\mj_d)\in \{0,1\}^d$$ 
let
    $$\mj-\mi=(\mj_1-\mi_1,\ldots,\mj_d-\mi_d)\in \{-1,0,1\}^d$$ 
stand for their usual coordinate-wise difference.
Let 
$\me^{(\ell)}$ stand for the multi-index with the only non-zero coordinate the $\ell$--th one, which is equal to 1.
\newcommand{\rel}{\prec}
For each $\ell=1,\ldots,\ell$ we define a relation on $\{0,1\}^d$ by
    $$
        \mi \rel_\ell \mj 
        \quad \Leftrightarrow \quad
        \mj-\mi=\me^{(\ell)}.
    $$

For a point $\x=(x_1,\ldots,x_d)\in \RR^d$ we define the functions
\begin{align*}
    G_d:\RR^d\to \RR,\quad 
    G_d(\x)
    &:=\sum_{i=1}^d x_i-d+1,\\
    H_d:\RR^d\to \RR,\quad
    H_d(\x)
    &:=\min\{x_1, x_2, \ldots, x_d\}.
\end{align*}
\smallskip

In our previous work 
we proved the following proposition.

\begin{proposition}[{\cite[Propositions 3.1 and 3.2]{BeOmVuZa}}]
\label{prel:prop}
Define the following linear program
\begin{align}
\label{LP}
\begin{split}
\min_{
\substack{
    a_1,\ldots,a_d,\\
    b_1,\ldots,b_d,\\
    q_{\mi} \text{ for }\mi\in \{0,1\}^d
}}
&\hspace{0.2cm} \sum_{\mi\in \{0,1\}^d} \sign(\mi) q_\mi,\\
\text{subject to }
&\hspace{0.2cm} 
    0\leq a_i< b_i\leq 1\quad i=1,\ldots,d,\\
&\hspace{0.2cm} 
    0\le q_{\mj} - q_{\mi} \le b_\ell - a_\ell
\quad 
    \text{for all }\ell=1,\ldots,d
    \text{ and all }\mi\rel_\ell\mj,\\
&\hspace{0.2cm} 
    \max\{0,G_d(x_{\mi})\} 
    \le q_\mi \le 
    H_d(x_{\mi})
    \quad \text{for all }\mi\in \{0,1\}^d.
\end{split}
\end{align}
Let $\mi^{(1)},\ldots,\mi^{(2^d)}$ be some order of all multi-indices $\mi\in\{0,1\}^d$.
    If there exists an optimal solution 
    $(a_1^\ast,\ldots,a_d^\ast,
    b_1^\ast,\ldots,b_d^\ast,
    q_{\mi^{(1)}},\ldots,q_{\mi^{(2^d)}}
    )$
    to \eqref{LP}, which
    satisfies 
    $        b_1^\ast=\ldots=b_d^\ast=1,$
    then the optimal value of \eqref{LP} is the maximal negative volume of some box over all $d$-quasi-copulas.

    Moreover, if there exists an optimal solution 
    to \eqref{LP} where $\min$ is replaced with $\max$, which
    satisfies 
    $        b_1^\ast=\ldots=b_d^\ast=1,$
    then the optimal value of \eqref{LP} is the maximal positive volume of some box over all $d$-quasi-copulas.
\end{proposition}

\begin{rem}
\label{rem:LP}
In \cite[Section 3]{BeOmVuZa} we presented numerical solutions to \eqref{LP}
up to $d=18$ for $\min$ and up to $d=17$ for $\max$. In the case of $\min$
and $7\leq d\leq 18$ we obtained optimal solutions with $b_1^\ast=\ldots=b_d^\ast=1$
and so we solved the maximal negative volume problem.
Cases $d< 7$ have to be considered separately by extending Proposition \ref{prel:prop}
from the $2^d$-element grid $\prod_{i=1}^d \{a_i,b_i\}$ to the $3^d$-element grid $\prod_{i=1}^d\{a_i,b_i,1\}$.
In the case of $\max$ 
and $2\leq d\leq 17$ we obtained optimal solutions with $b_1^\ast=\ldots=b_d^\ast=1$
and so we solved the maximal positive volume problem.
\hfill$\blacksquare$
\end{rem}


\subsection{Symmetrization of the linear program \eqref{LP} and the dual}
\label{symmetrization}

By the following proposition it suffices to consider symmetric solutions
to the linear program \eqref{LP}.

\begin{proposition}
\label{symmetric-solution}
Assume the notation from Proposition \ref{prel:prop}.
There exists an optimal solution to the linear program \eqref{LP} 
of the form
$$
    \big(
    \underbrace{a^\ast,\ldots,a^{\ast}}_{d},
    \underbrace{b^\ast,\ldots,b^{\ast}}_{d},
    q_{\|\mi^{(1)}\|_1}^\ast,\ldots,q_{\|\mi^{(2^d)}\|_1}^\ast
    \big)
$$
for some $a,b,q_1,\ldots,q_d\in [0,1]$.

    Analogously, replacing $\min$ with $\max$ in \eqref{LP} above,
    the same statement holds.
\end{proposition}

\begin{proof}
    Let $S_d$ be the set of all permutations of a $d$-element set
    $\{1,\ldots,d\}$.
    For $\Phi\in S_d$ and $\mi^{(j)}:=(\mi_1^{(j)},\ldots,\mi_d^{(j)})\in\{0,1\}^d$,
    let $\Phi(\mi^{(j)}):=\big(\mi_{\Phi(1)}^{(j)},\ldots,\mi_{\Phi(d)}^{(j)}\big)$.
    If 
    \begin{equation}
    \label{one-nonsymmetric-solution}
    \big(
    a_1^\ast,\ldots,a_d^{\ast},
    b_1^\ast,\ldots,b_d^{\ast},
    q_{\mi^{(1)}}^\ast,\ldots,q_{\mi^{(2^d)}}^\ast
    \big)
    \end{equation}
    is an optimal solution to \eqref{LP}, then
    \begin{equation}
    \label{second-nonsymmetric-solution}
    \big(
    a_{\Phi(1)}^\ast,\ldots,a_{\Phi(d)}^{\ast},
    b_{\Phi(1)}^\ast,\ldots,b_{\Phi(d)}^{\ast},
    q_{\Phi(\mi^{(1)})}^\ast,\ldots,q_{\Phi(\mi^{(2^d)})}^\ast
    \big)
    \end{equation}
    is also an optimal solution to \eqref{LP}.
    Hence, an optimal solution as stated in the proposition is equal to
    \begin{equation}
    \label{average}
    \frac{1}{n!}
    \sum_{\Phi\in S_n}
    \big(
    a_{\Phi(1)}^\ast,\ldots,a_{\Phi(d)}^{\ast},
    b_{\Phi(1)}^\ast,\ldots,b_{\Phi(d)}^{\ast},
    q_{\Phi(\mi^{(1)})}^\ast,\ldots,q_{\Phi(\mi^{(2^d)})}^\ast
    \big),
    \end{equation}
    where the summation is the coordinate-wise one.
    The proof for $\max$ instead of $\min$ is the same.
\end{proof}

An immediate corollary to Proposition \ref{symmetric-solution} is the following.

\begin{corollary}
\label{cor:sim}
    The optimal value of the linear program \eqref{LP} is equal to the optimal value of the 
    linear program
\begin{align}
\label{LP-symmetric}
\begin{split}
\min_{
\substack{
    a,
    b,
    q_0,q_{1},\ldots,q_d
}}
&\hspace{0.5cm} \sum_{i=0}^d (-1)^{d-i} \binom{d}{i} q_i,\\
\text{subject to }
&\hspace{0.5cm} 
    0\leq a< b\leq 1,\\
&\hspace{0.5cm} 
     0\le q_{i} - q_{i-1} \le b - a\quad \text{for }i=1,\ldots,d,\\
&\hspace{0.5cm} 
    \max\{0,(d-i)a+ib-d+1\} 
    \le q_i \le 
    a\\
    &\hspace{3cm}\text{for } i=0,\ldots,d-1,\\
&\hspace{0.5cm} 
    \max\{0,db-d+1\} 
    \le q_d \le 
    b.    
\end{split}
\end{align}

    Analogously, replacing $\min$ with $\max$ in \eqref{LP} above,
    the same statement holds.
\end{corollary}

It turns out that some of the constraints in \eqref{LP-symmetric} are redundant, while introducing new variables $\delta_i=q_i-q_{i-1}$
further decreases the number of constraints.

\begin{proposition}
\label{simplify-further}
    The optimal value of the linear program \eqref{LP-symmetric} is equal to the optimal value of the linear program
\begin{align}
\label{LP-symmetric-v3}
\begin{split}
\min_{
\substack{
    a,
    b,
    q_0,\delta_{1},\ldots,\delta_d
}}
&\hspace{0.5cm}
\sum_{j=1}^d (-1)^{d+j}\binom{d-1}{j-1}\delta_j,\\
\text{subject to }
&\hspace{0.5cm} 
    b\leq 1,\\
&\hspace{0.5cm} 
    \delta_i\le b - a\quad \text{for }i=1,\ldots,d,\\
&\hspace{0.5cm} 
    q_0+\sum_{i=1}^{d-1}\delta_i \le 
    a,\\
&\hspace{0.5cm} 
    db-d+1 
    \le q_0+\sum_{i=1}^d \delta_i,\\
&\hspace{0.5cm} 
    a\geq 0,\; b\geq 0,\; q_0\geq 0,\;\delta_i\geq 0 \quad \text{for }i=1,\ldots,d.
\end{split}
\end{align}

    Analogously, replacing $\min$ with $\max$ in \eqref{LP-symmetric-v3} above,
    the same statement holds.
\end{proposition}

\begin{proof}
  First let us prove that in \eqref{LP-symmetric} the conditions
    \begin{align}
    \label{redundant-conditions}
    \begin{split}
        &(d-i)a+ib-d+1\leq q_i \quad \text{for }i=0,\ldots,d-1,\\
        &q_i\leq a\quad \text{for }i=0,\ldots,d-2,\\
        &q_d\leq b
    \end{split}
    \end{align}
    follow from the other conditions.
    We have that
    \begin{align*}
        q_{d-1}=q_d+(q_{d-1}-q_d)\geq db-d+1+a-b=a+(d-1)b-d+1,
    \end{align*}
    where we used the second and the fourth conditions from 
    \eqref{LP-symmetric}
    for the inequality. This is the condition in the first line of \eqref{redundant-conditions} for
    $i=d-1$. Inductively, using 
        $$q_{d-j}=q_{d-j+1}+(q_{d-j}-q_{d+j+1}),$$ 
    the conditions in the first line of \eqref{redundant-conditions} for
    $i=d-2,\ldots,1$ follow.

    The conditions in the second line of \eqref{redundant-conditions} follow
    easily from 
        $$a\geq q_{d-1}\geq q_{d-2}\geq \ldots\geq q_0.$$
        
    Finally, the condition in the last line of \eqref{redundant-conditions} follows from
    $$
    q_d=q_{d-1}+(q_{d}-q_{d-1})
    \leq a+(b-a)=b,
    $$
    where we used the second and the third conditions from \eqref{LP-symmetric} for the inequality. 

    Now we introduce new variables $\delta_i=q_i-q_{i-1}$, $i=1,\ldots,d$
    and the constraints of the reduced linear program \eqref{LP-symmetric},
    i.e., \eqref{LP-symmetric} without the constraints from \eqref{redundant-conditions},
    become as in \eqref{LP-symmetric-v3}. 
      The only non-trivial thing to verify is the form of the objective function. But this follows by a simple computation:
    \begin{align*}
    \sum_{i=0}^d(-1)^{d-i}\binom{d}{i}q_i 
    &=  \sum_{i=0}^d(-1)^{d-i}\binom{d}{i}\Big(q_0 + \sum_{j=1}^i \delta_j\Big)  \\
    &= q_0\underbrace{\sum_{k=0}^d (-1)^{d-i}\binom{d}{i}}_{0} +
    \sum_{j=1}^d\delta_j 
        \underbrace{\Big(\sum_{k=j}^d (-1)^{d-k}\binom{d}{k}\Big)}_{
        (-1)^{d+j}\binom{d-1}{j-1}
        },
\end{align*}
where we used 
    $\sum_{k\leq n}(-1)^k\binom{r}{k}=(-1)^n\binom{r-1}{n}$
in the last equality.
\end{proof}

\subsection{The dual linear program of \eqref{LP-symmetric-v3}}
Recall that writing the linear program in the form
$
    \displaystyle\max_{x}\{c^Tx\colon Ax\le b, x\ge 0\}
$
where $c\in \RR^n$ and $b\in \RR^m$ are real vectors, 
$A\in \RR^{m\times n}$ is a matrix,
$x$ is a vector of variables and the inequalities are the coordinate-wise ones,
its dual program is given by
$
    \displaystyle\min_{y}\{ b^Ty\colon A^Ty\ge c, y\ge 0\},
$
where $y$ is a vector of dual variables.
Namely, for each constraint of the primal program, there is a variable for the dual program,
while for each variable of the primal program, there is a constraint for the dual program.

Rewriting the objective function $\displaystyle \min_x c^T x$ of \eqref{LP-symmetric-v3} in
the form $\displaystyle -(\max_x(-c)^T x)$
the dual linear program of \eqref{LP-symmetric-v3} is the following:

\begin{align}
\label{dual-LP-symmetric-v3}
\begin{split}
-\Big(\min_{
\substack{
    y_1,
    l_1,\ldots,l_d,
    y_2,y_3
}}
&\hspace{0.5cm} y_1+(d-1)y_3\Big),\\
\text{subject to }
&\hspace{0.5cm} 
    \sum_{i=1}^d l_i  - y_2 \ge 0,\\
&\hspace{0.5cm}
    y_1 -\sum_{i=1}^d l_i + dy_3 \ge 0,\\
&\hspace{0.5cm}
    y_2 - y_3\ge 0,\\ 
&\hspace{0.5cm}
    l_j+y_2-y_3\ge (-1)^{d+j+1}\binom{d-1}{j-1}\quad
    \text{for }j=1,\ldots,d-1,\\[0.3em]
&\hspace{0.5cm}
    l_d - y_3\geq -1,\\
&\hspace{0.5cm} 
    y_1\geq 0,\; y_2\geq 0,\;y_3\geq 0,\; 
    l_i\geq 0 \quad \text{for }i=1,\ldots,d.
\end{split}
\end{align}

Analogously, the dual linear program of \eqref{LP-symmetric-v3} 
with the objective function $\displaystyle \max_x c^Tx$ is the following:

\begin{align}
\label{dual-LP-symmetric-v3-max}
\begin{split}
\min_{
\substack{
    y_1,
    l_1,\ldots,l_d,
    y_2,y_3
}}
&\hspace{0.5cm} y_1+(d-1)y_3,\\
\text{subject to }
&\hspace{0.5cm} 
    \sum_{i=1}^d l_i  - y_2 \ge 0,\\
&\hspace{0.5cm}
    y_1 -\sum_{i=1}^d l_i + dy_3 \ge 0,\\
&\hspace{0.5cm}
    y_2 - y_3\ge 0,\\ 
&\hspace{0.5cm}
    l_j+y_2-y_3\ge (-1)^{d+j}\binom{d-1}{j-1}\quad
    \text{for }j=1,\ldots,d-1,\\[0.3em]
&\hspace{0.5cm}
    l_d - y_3\geq 1,\\
&\hspace{0.5cm} 
    y_1\geq 0,\; y_2\geq 0,\;y_3\geq 0,\; 
    l_i\geq 0 \quad \text{for }i=1,\ldots,d.
\end{split}
\end{align}

It turns out that both linear programs can be simplified.

\begin{proposition}
\label{prop:simplification}
The optimal solutions to \eqref{dual-LP-symmetric-v3}
and \eqref{dual-LP-symmetric-v3-max}
are equal to the optimal solutions to 
\begin{align}
\label{dual-LP-symmetric-v4-new}
\begin{split}
-\Big(\min_{
\substack{
    l_1,\ldots,l_d,
    y_3
}}
&\hspace{0.5cm} (d-1)y_3\Big),\\
\text{subject to }
&\hspace{0.5cm}
    -\sum_{i=1}^d l_i + dy_3 = 0,\\ 
&\hspace{0.5cm}
    l_j+(d-1)y_3\ge (-1)^{d+j+1}\binom{d-1}{j-1}\quad
    \text{for }j=1,\ldots,d-1,\\[0.3em]
&\hspace{0.5cm}
    l_d=\max(0, -1+y_3),\\
&\hspace{0.5cm} 
    y_3\geq 0,\; 
    l_i\geq 0 \quad \text{for }i=1,\ldots,d,
\end{split}
\end{align}
and
\begin{align}
\label{dual-LP-symmetric-v4-max-new}
\begin{split}
\min_{
\substack{
    l_1,\ldots,l_d,
    y_3
}}
&\hspace{0.5cm} (d-1)y_3,\\
\text{subject to }
&\hspace{0.5cm}
    -\sum_{i=1}^d l_i + dy_3 = 0,\\
&\hspace{0.5cm}
    l_j+ (d-1)y_3\ge (-1)^{d+j}\binom{d-1}{j-1}\quad
    \text{for }j=1,\ldots,d-1,\\[0.3em]
&\hspace{0.5cm}
    l_d - y_3=1,\\
&\hspace{0.5cm} 
    y_3\geq 0,\; 
    l_i\geq 0 \quad \text{for }i=1,\ldots,d,
\end{split}
\end{align}
respectively.
\end{proposition}

\begin{proof}
We will prove the proposition by first establishing a few claims.\\

\noindent \textbf{Claim 1.} Their exists an optimal solution $(y^{\opt}_1,l^{\opt}_1,\ldots,l^{\opt}_d,y^{\opt}_2,y^{\opt}_3)$ 
to the linear program 
    \eqref{dual-LP-symmetric-v3} (resp.\
    \eqref{dual-LP-symmetric-v3-max})
which
satisfies
    $y_2^{\opt}=y_1^{\opt}+dy_3^{\opt}.$\\

\noindent \textit{Proof of Claim 1.}
Let
    $(y^{\opt}_1,l^{\opt}_1,\ldots,l^{\opt}_d,y^{\opt}_2,y^{\opt}_3)$ 
be an optimal solution to \eqref{dual-LP-symmetric-v3}.
From the first two constraints in \eqref{dual-LP-symmetric-v3} 
it follows that
\begin{equation}
\label{ineq-for-y2opt}
    y_1^{\opt}+dy_3^{\opt}\geq \sum_{i=1}^d l_i \geq y_2^{\opt}.
\end{equation}

If the left inequality in \eqref{ineq-for-y2opt} is strict and $y^{\ast}_1>0$, then there exists
$\epsilon>0$ such that 
    $(y^{\opt}_1-\epsilon,l^{\opt}_1,\ldots,l^{\opt}_d,y^{\opt}_2,y^{\opt}_3)$
is a feasible solution to \eqref{dual-LP-symmetric-v3}
such that
    $$(y_1^\ast-\epsilon)+(d-1)y_3^{\ast}<y_1^\ast+(d-1)y_3^{\ast}.$$
But this contradicts to the optimality of 
$(y^{\opt}_1,l^{\opt}_1,\ldots,l^{\opt}_d,y^{\opt}_2,y^{\opt}_3)$.

If the left inequality in \eqref{ineq-for-y2opt} is strict, $y^{\ast}_1=0$ and $y^{\ast}_3>0$, then there exists
$\epsilon>0$ such that 
    $
    (y^{\opt}_1,l^{\opt}_1,\ldots,l^{\opt}_d,
    y^{\opt}_2-\epsilon,y^{\opt}_3-\epsilon)
    $
is a feasible solution to \eqref{dual-LP-symmetric-v3}
such that
$$y_1^\ast+(d-1)(y_3^{\ast}-\epsilon)<y_1^\ast+(d-1)y_3^{\ast}.$$
But this contradicts to the optimality of 
$(y^{\opt}_1,l^{\opt}_1,\ldots,l^{\opt}_d,y^{\opt}_2,y^{\opt}_3)$.

Hence, the left inequality in \eqref{ineq-for-y2opt} is an equality.
If the right inequality is strict, we just increase $y_2^{\opt}$ to 
$\sum_{i=1}^d l_i$ and we get another optimal solution to \eqref{dual-LP-symmetric-v3}.
This proves the lemma for \eqref{dual-LP-symmetric-v3}.

The proof for \eqref{dual-LP-symmetric-v3-max} is the same. 
\hfill$\blacksquare$\\

Using Claim 1,
the linear programs \eqref{dual-LP-symmetric-v3} and
\eqref{dual-LP-symmetric-v3-max} simplify to:

\begin{align}
\label{dual-LP-symmetric-v4}
\begin{split}
-\Big(\min_{
\substack{
    y_1,
    l_1,\ldots,l_d,
    y_3
}}
&\hspace{0.5cm} y_1+(d-1)y_3\Big),\\
\text{subject to }
&\hspace{0.5cm}
    y_1 -\sum_{i=1}^d l_i + dy_3 = 0,\\
&\hspace{0.5cm}
    y_1 + (d-1)y_3\ge 0,\\ 
&\hspace{0.5cm}
    l_j+y_1+ (d-1)y_3\ge (-1)^{d+j+1}\binom{d-1}{j-1}\\[0.3em]
&\hspace{3cm}
    \text{for }j=1,\ldots,d-1,\\
&\hspace{0.5cm}
    l_d - y_3\geq -1,\\
&\hspace{0.5cm} 
    y_1\geq 0,\; \;y_3\geq 0,\; 
    l_i\geq 0 \quad \text{for }i=1,\ldots,d,
\end{split}
\end{align}

and

\begin{align}
\label{dual-LP-symmetric-v4-max}
\begin{split}
\min_{
\substack{
    y_1,
    l_1,\ldots,l_d,
    y_3
}}
&\hspace{0.5cm} y_1+(d-1)y_3,\\
\text{subject to }
&\hspace{0.5cm}
    y_1 -\sum_{i=1}^d l_i + dy_3 = 0,\\
&\hspace{0.5cm}
    y_1+ (d-1)y_3\ge 0,\\ 
&\hspace{0.5cm}
    l_j+y_1+ (d-1)y_3\ge (-1)^{d+j}\binom{d-1}{j-1}\\[0.3em]
&\hspace{3cm}
    \text{for }j=1,\ldots,d-1,\\
&\hspace{0.5cm}
    l_d - y_3\geq 1,\\
&\hspace{0.5cm} 
    y_1\geq 0,\; \;y_3\geq 0,\; 
    l_i\geq 0 \quad \text{for }i=1,\ldots,d.
\end{split}
\end{align}
\smallskip

\noindent \textbf{Observation 1.} Since $y_1\geq 0, y_3\geq 0$, the inequality 
$y_1+(d-1)y_3\geq 0$
in \eqref{dual-LP-symmetric-v4} and \eqref{dual-LP-symmetric-v4-max}
is redundant.\\

\noindent \textbf{Claim 2.}
There exists an optimal solution $(y^{\opt}_1,l^{\opt}_1,\ldots,l^{\opt}_d,y^{\opt}_3)$ 
to the linear program 
    \eqref{dual-LP-symmetric-v4} (resp.\
    \eqref{dual-LP-symmetric-v4-max})
which
satisfies
    $y_1^{\opt}=0.$\\

\noindent\textit{Proof of Claim 2.}
Let
    $(y^{\opt}_1,l^{\opt}_1,\ldots,l^{\opt}_d,y^{\opt}_2,y^{\opt}_3)$ 
be an optimal solution to \eqref{dual-LP-symmetric-v4}.
Assume that $y_1^{\opt}>0$. Then
 $
 \big(0,l^{\opt}_1,\ldots,l^{\opt}_{d-1},
 l^{\opt}_{d}+\frac{y_1^\ast}{d-1},
 y^{\opt}_3+\frac{y_1^\ast}{d-1}\big)
 $
is another optimal solution to \eqref{dual-LP-symmetric-v4}.

The proof for \eqref{dual-LP-symmetric-v4-max} is the same.
\hfill$\blacksquare$\\

Using Observation 1
and Claim 2,
the linear programs \eqref{dual-LP-symmetric-v4} and
\eqref{dual-LP-symmetric-v4-max} simplify to 
\begin{align}
\label{dual-LP-symmetric-v5}
\begin{split}
-\Big(\min_{
    l_1,\ldots,l_d,
    y_3}
&\hspace{0.5cm} (d-1)y_3\Big),\\
\text{subject to }
&\hspace{0.5cm}
    -\sum_{i=1}^d l_i + dy_3 = 0,\\
&\hspace{0.5cm}
    l_j+(d-1)y_3\ge (-1)^{d+j+1}\binom{d-1}{j-1}\quad
    \text{for }j=1,\ldots,d-1,\\[0.3em]
&\hspace{0.5cm}
    l_d - y_3\geq -1,\\
&\hspace{0.5cm} 
    y_3\geq 0,\; 
    l_i\geq 0 \quad \text{for }i=1,\ldots,d,
\end{split}
\end{align}
and
\begin{align}
\label{dual-LP-symmetric-v5-max}
\begin{split}
\min_{
    l_1,\ldots,l_d,
    y_3}
&\hspace{0.5cm} (d-1)y_3,\\
\text{subject to }
&\hspace{0.5cm}
    -\sum_{i=1}^d l_i + dy_3 = 0,\\ 
&\hspace{0.5cm}
    l_j+(d-1)y_3\ge (-1)^{d+j}\binom{d-1}{j-1}\quad
    \text{for }j=1,\ldots,d-1,\\[0.3em]
&\hspace{0.5cm}
    l_d - y_3\geq 1,\\
&\hspace{0.5cm} 
    y_3\geq 0,\; 
    l_i\geq 0 \quad \text{for }i=1,\ldots,d,
\end{split}
\end{align}
respectively.\\

The following claim shows that in the optimal solution to the linear programs  
    \eqref{dual-LP-symmetric-v5} 
and 
    \eqref{dual-LP-symmetric-v5-max}, 
    one of constraints $l_d\geq 0$ and $l_d-y_3\geq \pm 1$ is an equality. \\

\noindent \textbf{Claim 3.}
If $(y^{\opt}_1,l^{\opt}_1,\ldots,l^{\opt}_d,y^{\opt}_3)$ 
is an optimal solution
to the linear program 
    \eqref{dual-LP-symmetric-v5} (resp.\
    \eqref{dual-LP-symmetric-v5-max}),
then
    $
    l_d^{\opt}=\max(0,-1+y_3^{\opt})$
    $(\text{resp. }l_d^{\opt}=1+y_3^{\opt})$.
        
Moreover, $l_d^\ast=0$ can only occur for $d\leq 6$.\\

\noindent \textit{Proof of Claim 3.}
Let $(y^{\opt}_1,l^{\opt}_1,\ldots,l^{\opt}_d,y_3^\ast)$ be an optimal solution
to \eqref{dual-LP-symmetric-v5}.
Assume that $c:=l_d^{\opt}-\max(0,-1+y_3^{\opt})>0.$
Then there exists $\epsilon>0$ such that 
    \begin{equation*}
            (y_1^\ast,
            l_1^\ast+(d-1)\epsilon,
            \ldots,
            l_{d-1}^\ast+(d-1)\epsilon,
            l_d-(d^2-d+1)\epsilon,
            y_3^{\ast}-\epsilon)
    \end{equation*}
is still a feasible solution to \eqref{dual-LP-symmetric-v5} with a smaller objective function, which is a contradiction to the optimality of $(y^{\opt}_1,l^{\opt}_1,\ldots,l^{\opt}_d,y_3^\ast)$.

The proof for \eqref{dual-LP-symmetric-v5-max} is the same, 
using also  that $l_d^\ast=\max(0,1+y_3^\ast)=1+y_3^\ast$.

Let us establish the moreover part. In this case $y_3^{\opt}\leq 1$. 
Since every $y_3$ large enough extends to a feasible solution to \eqref{dual-LP-symmetric-v5}
(e.g., $l_1=\ldots=l_{d-1}=0$ and $l_d=dy_3$),
convexity of the feasible region implies that there is a feasible solution with $y_3=1$. But then for this solution the first constraint in \eqref{dual-LP-symmetric-v5} implies that 
\begin{equation}
    \label{equality}
        \sum_{i=1}^d l_i=d. 
\end{equation}

We separate two cases according to the parity of $d$.\\

\noindent \textbf{Case 1:} $d=2d'$ for some $d'\in \NN$. Summing up the inequalities with positive right hand side in 
\eqref{dual-LP-symmetric-v5} we get 
\begin{equation}
\label{inequality}
    \sum_{\substack{j\text{ odd},\\ j<d}} l_{j}+(d-1)\Big(\frac{d}{2}-1\Big)
    \geq 
    \sum_{\substack{j \text{ odd},\\ j<d}} \binom{d-1}{j-1}.
\end{equation}
Using \eqref{equality} in \eqref{inequality}, it follows that
\begin{align*}
d+(d-1)\Big(\frac{d}{2}-1\Big)
\geq 
\sum_{\substack{j \text{ odd},\\ j<d}} \binom{d-1}{j-1}
&=\binom{d-1}{0}+\binom{d-1}{2}+\ldots+\binom{d-1}{d-2}.
\end{align*}
For $d\geq 8$ this is a contradiction.\\

\noindent \textbf{Case 2:} $d=2d'-1$ for some $d'\in \NN$. Summing up the inequalities with positive right hand side in 
\eqref{dual-LP-symmetric-v5} above we get 
\begin{equation}
\label{inequality-2}
\sum_{\substack{j\text{ even},\\ j<d}} l_{j}
+\frac{(d-1)^2}{2}
\geq 
\sum_{\substack{j\text{ even},\\ j<d}}\binom{d-1}{j-1}.
\end{equation}
Using \eqref{equality} in \eqref{inequality-2}, it follows that
$$
d+\frac{(d-1)^2}{2}
\geq 
\sum_{\substack{j \text{ even},\\ j<d}}
\binom{d-1}{j-1}
=\binom{d-1}{1}+\binom{d-1}{3}+\ldots+\binom{d-1}{d-2}
.$$
For $d\geq 7$ this is a contradiction.

This proves Claim 3.\hfill$\blacksquare$\\

Using 
Claim 3,
the linear programs \eqref{dual-LP-symmetric-v5} and
\eqref{dual-LP-symmetric-v5-max} simplify to the linear programs
\eqref{dual-LP-symmetric-v4-new} and \eqref{dual-LP-symmetric-v4-max-new},
respectively.
\end{proof}


\section{Basic technical result}
\label{sec:basic-technical}

The following proposition is the basic result, which will be used to prove Theorems \ref{sol:min-value} and \ref{sol:max-value}. 

\begin{proposition}
    \label{auxiliary-proposition}
    Let 
        $$0<c_1\leq c_2\leq \ldots\leq c_k$$ 
    be given positive real numbers with $c_1<c_k$, 
        $$e_1<0,\ldots,e_r<0$$ 
    given negative real numbers,
    $\alpha\in \RR$ such that 
        $$-c_k<\alpha\leq c_k$$
    and a linear program
\begin{align}
\label{dual-LP-simplified}
\begin{split}
\min_{
\substack{
w,y_1,y_2,\ldots,y_k,\\[0.2em]
z_1,z_2,\ldots,z_r
}} 
&\hspace{1cm} w\\
\text{subject to }
&\hspace{1cm} 
    y_1+\ldots+y_k+z_1+\ldots+z_r=w+\alpha,\\
&\hspace{1cm}
    y_i+w\geq c_i
\quad 
    \text{for }i=1,\ldots,k,\\
&\hspace{1cm}
    z_i+w\geq e_i
\quad 
    \text{for }i=1,\ldots,r,\\
&\hspace{1cm}
    w\geq 0,\\
&\hspace{1cm}
    y_i\geq 0\quad \text{for }i=1,\ldots,k,\\
&\hspace{1cm}
    z_i\geq 0\quad \text{for }i=1,\ldots,r.
\end{split}
\end{align}
Define 
    $$\displaystyle w_i:=\frac{1}{i+1}\Big(\sum_{j=0}^{i-1} c_{k-j} -\alpha\Big) \quad \text{for }i=1,\ldots,k.$$
and $c_0:=0$. Let 
\begin{equation}
\label{def:i0}
    i_0
    \text{ be the smallest integer in } \{1,\ldots,k\}
    \text{ such that }
    w_{i_0}\geq c_{k-i_0}.
\end{equation}
Then the optimal solution 
$(w^{\ast},y_1^\ast,\ldots,y^\ast_k,z_1^\ast,\ldots,z_r^\ast)$ 
to \eqref{dual-LP-simplified} is
    $$
        (w^{\ast},y_1^\ast,\ldots,y^\ast_k,z_1^\ast,\ldots,z_r^\ast)
        =
        (w_{i_0},\underbrace{0,\ldots,0}_{k-i_0},c_{k-i_0+1}-w_{i_0},\ldots,c_{k}-w_{i_0},
        \underbrace{0,\ldots,0}_{r}).
    $$
\end{proposition}

\begin{proof}
    First we establish two claims.\\
    
    \noindent \textbf{Claim 1.} There is at most one $i\in \{1,\ldots,k\}$ such that 
        $w_i\in [c_{k-i},c_{k-i+1})$.\\

    \noindent\textit{Proof of Claim 1.} 
    Let $\widetilde i$ be such that $w_{\widetilde i}\in [c_{k-\widetilde i},c_{k-\widetilde i+1})$. 
    Note that $w_{\widetilde i}\geq c_{k-\widetilde i}$ is equivalent to
    \begin{equation}
        \label{ineq-1}
            \sum_{j=0}^{\widetilde i-1} c_{k-j}-\alpha\geq (\widetilde i+1)c_{k-\widetilde i}.
    \end{equation}
    Hence, 
    $$
    w_{\widetilde i+1}
    =\frac{1}{\widetilde i+2}\Big(\sum_{j=0}^{\widetilde i} c_{k-j}-\alpha\Big)
    \underbrace{\geq}_{\eqref{ineq-1}}
    c_{k-\widetilde i},
    $$
    which in particular implies that $w_{\widetilde i+1}\notin [c_{k-\widetilde i-1},c_{k-\widetilde i})$.
    Inductively we can prove that 
    $
    w_{\widetilde i+l}\geq c_{k-\widetilde i}
    $
    for $l=1,2,\ldots,k-\widetilde i$ 
    ,
    whence
    $w_{\widetilde i+l}\notin [c_{k-\widetilde i-l},c_{k-\widetilde i-l+1})$.
    This proves Claim 1.\hfill$\blacksquare$\\

    \noindent \textbf{Claim 2.} Let $i_0$ be as in \eqref{def:i0}. Then
        $w_{i_0}\in [c_{k-i_0},c_{k-i_0+1}).$\\
        
    \noindent \textit{Proof of Claim 2.} 
    First note that $i_0$ is well-defined, since there exists 
    $i\in \{1,\ldots,k\}$ such that $w_i\geq c_{k-i}$, e.g., $i=k$
    due to $w_k=\frac{1}{k}\big(\sum_{j=1}^k c_k-\alpha\big)\geq 0=c_0.$
    Also note that $i=i_0$ is the unique candidate for the containment 
        $w_i\in [c_{k-i},c_{k-i+1})$.
    This follows from the following two observations:
    \begin{itemize}
        \item By definition of $i_0$, we have that $w_{i}<c_{k-i}$ 
            for $i=1,\ldots,i_0-1$.
        \smallskip
        \item As in the proof of Claim 1, $w_i\geq c_{k-i_0}\geq c_{k-i+1}$ for $i=i_0+1,\ldots,k$.
    \end{itemize}
    So it only remains to prove that $w_{i_0}<c_{k-i_0+1}$. Assume on the contrary that 
    $w_{i_0}\geq c_{k-i_0+1}$. First notice that in this case $i_0\neq 1$, since 
    $w_{1}=\frac{1}{2}(c_k-\alpha)< \frac{1}{2}(c_k+c_k)=c_k$. 
    Then
    \begin{equation*}
        \label{ineq-2}
            \sum_{j=0}^{i_0-1} c_{k-j}-\alpha\geq (i_0+1)c_{k-i_0+1},
    \end{equation*}
    which implies that
    \begin{equation*}
        \label{ineq-3}
            \sum_{j=0}^{i_0-2} c_{k-j}-\alpha\geq i_0\cdot c_{k-i_0+1}.
    \end{equation*}
    Further on, 
    $$
    w_{i_0-1}
    =\frac{1}{i_0}\Big(\sum_{j=0}^{i_0-2} c_{k-j}-\alpha\Big)
    \geq
    c_{k-i_0+1}.
    $$
    But this is a contradiction with the minimiality of $i_0$.\hfill$\blacksquare$\\

    Let now $(w^{\ast},y_1^\ast,\ldots,y^\ast_k,z_1^\ast,\ldots,z_r^\ast)$ be an optimal solution to \eqref{dual-LP-simplified}.\\

    \noindent \textbf{Claim 3.} $z_i^\ast=0$ for $i=1,\ldots,r$.\\

    \noindent \textit{Proof of Claim 3.}
    If there is $i$ such that $z_i>0$, then there exists $\epsilon>0$, 
    such that 
    \begin{equation}
        \label{also-feasible-v2}
            (w^{\ast}-\epsilon,
            y_1^\ast+\epsilon,
            \ldots,
            y_k^\ast+\epsilon,
            z_1^\ast,\ldots,
            z_{i-1}^\ast,
            z_{i}^\ast-(k+1)\epsilon,
            z_{i+1}^\ast,
            \ldots,
            z_r^\ast)
    \end{equation}
    is still a feasible solution
    to \eqref{dual-LP-simplified}. This contradicts to the optimality of $w^{\ast}$.
    \hfill $\blacksquare$\\
    
    Clearly $w=c_k$ extends to a feasible solution of \eqref{dual-LP-simplified}, e.g., 
        $y_1=c_k+\alpha$, 
        $y_i=0$ for $i=2,\ldots,k$
        and $z_i=0$ for $i=1,\ldots,r$. 
    So $w^{\ast}\leq c_k$.
    We separate two cases for the value of $w^\ast$.
    \\
    
    \noindent \textbf{Case 1.} $w^\ast=c_k$. Note that there is $i$ such that $y_i>0$.
    But then there exists $\epsilon>0$, such that 
    \begin{equation}
        \label{also-feasible}
            (w^{\ast}-\epsilon,
            y_1^\ast+\epsilon,
            \ldots,
            y_{i-1}^\ast+\epsilon,
            y_i^\ast-k\epsilon,
            y_{i+1}^\ast+\epsilon,
            \ldots,
            y_k^\ast+\epsilon,
            \underbrace{0,\ldots,0}_r)
    \end{equation}
    is still a feasible solution of \eqref{dual-LP-simplified}, which contradicts to the optimality of $w^{\ast}$.\\

    \noindent \textbf{Case 2.} $w^\ast<c_k$. 
    Clearly 
        $y_{i}^\ast\geq \max(c_{i}-w^{\ast},0)$. 
    If there exists $i$ such that $y_i^\ast>\max(c_{i}-w^{\ast},0)$,
    then there exists $\epsilon>0$, such that \eqref{also-feasible} is still a feasible solution
    to \eqref{dual-LP-simplified}. This contradicts to the optimality of $w^{\ast}$.
    So $y_{i}^\ast=\max(c_{i}-w^{\ast},0)$.
    Let $\widehat i_0$ be such that 
        $w^\ast\in [c_{k-\widehat i_0},c_{k-\widehat i_0+1})$.
    Then 
    $$
        y_1^\ast=\ldots=y^\ast_{k-\widehat i_0}=0\quad \text{and}\quad 
        y_{i}^\ast=c_{i}-w^{\ast} \text{ for }i>k-\widehat i_0.
    $$
    Further on,
        $$
        \sum_{i=1}^{k} y_i^\ast
        =\sum_{i=k-\widehat i_0+1}^k (c_{i}-w^{\ast})
        =w^\ast+\alpha,
        $$
    whence
        $$w^{\ast}
        =\frac{1}{\widehat i_0+1} \Big(\sum_{i=k-\widehat i_0+1}^k c_{i}-\alpha\Big)
        =\frac{1}{\widehat i_0+1} \Big(\sum_{j=0}^{\widehat i_0-1} c_{k-j}-\alpha\Big)
        =w_{\widehat i_0}
        .$$
    Since by Claims 1 and 2, the only $i$ such that $w_i\in [c_{k-i_0},c_{k-i_0+1})$ is $i_0$ 
    defined by \eqref{def:i0}, it follows that $\widehat i_0=i_0$ and the optimal solution to
    \eqref{dual-LP-simplified} is as stated in the
    proposition.
\end{proof}


\section{Final step in the proof of Theorem \ref{sol:min-value}}
\label{sec:minimal}

Assume the notation as in the statement of Theorem \ref{sol:min-value}
and Section \ref{symmetrization}.
In this section we will solve the linear programs
\eqref{dual-LP-symmetric-v4-new} and 
\eqref{LP-symmetric},
which 
by Proposition \ref{prel:prop} also 
solves the maximal negative volume quasi-copula problem and proves Theorem \ref{sol:min-value}.\\

Let $(l^{\opt}_1,\ldots,l^{\opt}_d,y^{\opt}_3)$ be an optimal solution
to \eqref{dual-LP-symmetric-v4-new}.
By the moreover part of Claim 3 in the proof of Proposition \ref{prop:simplification},
under the assumption $d\geq 7$,
we have that $l_d^{\opt}=-1+y^{\opt}_3$.
Writing $w:=(d-1)y_3$, \eqref{dual-LP-symmetric-v4-new} becomes
\begin{align}
\label{dual-LP-symmetric-v6-new}
\begin{split}
-\Big(\min_{
    l_1,\ldots,l_{d-1},
    w}
&\hspace{0.5cm} w\Big),\\
\text{subject to }
&\hspace{0.5cm}
    -\sum_{i=1}^{d-1} l_i + w+1= 0,\\
&\hspace{0.5cm}
    l_j+w\ge (-1)^{d+j+1}\binom{d-1}{j-1}\quad
    \text{for }j=1,\ldots,d-1,\\[0.3em]
&\hspace{0.5cm} 
    w\geq 0,\; 
    l_i\geq 0 \quad \text{for }i=1,\ldots,d-1.
\end{split}
\end{align}
The solution to \eqref{dual-LP-symmetric-v6-new} will lean on the use of Proposition \ref{auxiliary-proposition}. 
Let $c_i$, $i=1,\ldots,\lfloor \frac{d}{2}\rfloor$ 
be as in Theorem \ref{sol:min-value}.
Let $r:=d-1-\lfloor\frac{d}{2}\rfloor$ and let $e_1,\ldots,e_r$ be the right hand sides $-\binom{d-1}{j-1}$ in \eqref{dual-LP-symmetric-v6-new} in some order.
By Proposition \ref{auxiliary-proposition} with $\alpha=1$, the optimal solution 
to \eqref{dual-LP-symmetric-v6-new} is 
    $w^\ast=w_{i_0,-}$
and:
\begin{enumerate}
\item If $d$ is even, then
    \begin{align}
    \label{optimal-v1}
    l_{j}^\ast&=
        \left\{
        \begin{array}{rl}
            \binom{d-1}{j-1}-w_{i_0,-},& 
                j\text{ is odd and }\binom{d-1}{j-1}>c_{\frac{d}{2}-i_0},\\[0.3em]
            0,& \text{otherwise}.
        \end{array}
        \right.
    \end{align}
\item If $d$ is odd, then
    \begin{align}
    \label{optimal-v2}
    l_{j}^\ast&=
        \left\{
        \begin{array}{rl}
            \binom{d-1}{j-1}-w_{i_0,-},& 
                j\text{ is even and }\binom{d-1}{j-1}>  
                    c_{\lfloor\frac{d}{2}\rfloor-i_0},\\[0.3em]
            0,& \text{otherwise}.
        \end{array}
        \right.
    \end{align}
\end{enumerate}
Hence, in an optimal solution 
to \eqref{dual-LP-symmetric-v3}
except \eqref{optimal-v1} or \eqref{optimal-v2},
also the following hold:
    $y_1^\ast=0$,
    $l_d^{\ast}=-1+\frac{1}{d-1}w_{i_0,-}$,
    $y_2^\ast=\frac{d}{d-1}w_{i_0,-}$,
    $y_3^\ast=\frac{1}{d-1}w_{i_0,-}$.
Using the complementary slackness, in an optimal solution 
to the dual \eqref{LP-symmetric-v3} of \eqref{dual-LP-symmetric-v3}, we have
    $a^\ast=\frac{i_0}{i_0+1}$,
    $b^{\ast}=1$,
    $q_0^\ast=0$
    and 
    $\delta_j^\ast$ are as in 
    \eqref{delta-min-optimal-even} if $d$ is even,
    while 
    $\delta_j^\ast$ are as in 
    \eqref{delta-min-optimal-odd} if $d$ is odd.
    This give the desired solution to \eqref{LP-symmetric}
    and proves Theorem \ref{sol:min-value}.



\section{Final step in the proof of Theorem \ref{sol:max-value}}
\label{sec:maximal}

Assume the notation as in the statement of Theorem \ref{sol:max-value}
and Section \ref{symmetrization}.
In this section we will solve the linear program
\eqref{dual-LP-symmetric-v4-max-new} and \eqref{LP-symmetric}, where $\min$ is replaced by $\max$,
This solves the maximal positive volume quasi-copula problem and proves Theorem \ref{sol:max-value}.\\

Let $(l^{\opt}_1,\ldots,l^{\opt}_d,y^{\opt}_3)$ be an optimal solution
to \eqref{dual-LP-symmetric-v4-max-new}.
By the moreover part of Claim 3 in the proof of Proposition \ref{prop:simplification}, 
we have that
    $l_d^{\opt}=1+y^{\opt}_3.$
Writing $w:=(d-1)y_3$, \eqref{dual-LP-symmetric-v5-max} becomes

\begin{align}
\label{dual-LP-symmetric-v6-max}
\begin{split}
\min_{
    l_1,\ldots,l_{d-1},
    w}
&\hspace{0.5cm} w,\\
\text{subject to }
&\hspace{0.5cm}
    -\sum_{i=1}^{d-1} l_i + w-1= 0,\\
&\hspace{0.5cm}
    l_j+w\ge (-1)^{d+j}\binom{d-1}{j-1}\quad
    \text{for }j=1,\ldots,d-1,\\[0.3em]
&\hspace{0.5cm} 
    w\geq 0,\; 
    l_i\geq 0 \quad \text{for }i=1,\ldots,d-1.
\end{split}
\end{align}
The solution to \eqref{dual-LP-symmetric-v6-max} will lean on the use of Proposition \ref{auxiliary-proposition}. 
Let $c_i$, $i=1,\ldots,\lfloor \frac{d+1}{2}\rfloor-1$ 
be as in Theorem \ref{sol:max-value}.
Let $r:=d-\lfloor \frac{d+1}{2}\rfloor$ and let $e_1,\ldots,e_r$ be the right hand sides $-\binom{d-1}{j-1}$ in \eqref{dual-LP-symmetric-v6-max} in some order.
By Proposition \ref{auxiliary-proposition} with $\alpha=-1$, the optimal solution 
to \eqref{dual-LP-symmetric-v6-max} is 
    $w^\ast=w_{i_0,+}$
and:
\begin{enumerate}
\item If $d$ is even, then
    \begin{align}
    \label{optimal-v1-max}
    l_{j}^\ast&=
        \left\{
        \begin{array}{rl}
            \binom{d-1}{j-1}-w_{i_0,-},& 
                j\text{ is even and }\binom{d-1}{j-1}>c_{\frac{d}{2}-1-i_0},\\[0.3em]
            0,& \text{otherwise}.
        \end{array}
        \right.
    \end{align}
\item If $d$ is odd, then
    \begin{align}
    \label{optimal-v2-max}
    l_{j}^\ast&=
        \left\{
        \begin{array}{rl}
            \binom{d-1}{j-1}-w_{i_0,-},& 
                j\text{ is odd and }\binom{d-1}{j-1}>  
                    c_{\lfloor\frac{d+1}{2}\rfloor-1-i_0},\\[0.3em]
            0,& \text{otherwise}.
        \end{array}
        \right.
    \end{align}
\end{enumerate} 
Hence, in an optimal solution 
to \eqref{dual-LP-symmetric-v3-max}
except \eqref{optimal-v1-max} or \eqref{optimal-v2-max},
also the following hold:
    $y_1^\ast=0$,
    $l_d^{\ast}=1+\frac{1}{d-1}w_{i_0,+}$,
    $y_2^\ast=\frac{d}{d-1}w_{i_0,+}$,
    $y_3^\ast=\frac{1}{d-1}w_{i_0,+}$.
Using the complementary slackness, in an optimal solution 
to the dual \eqref{LP-symmetric-v3}, with $\max$ instead of $\min$, of \eqref{dual-LP-symmetric-v3-max}, we have
    $a^\ast=\frac{i_0}{i_0+1}$,
    $b^{\ast}=1$,
    $q_0^\ast=0$
    and 
    $\delta_j^\ast$ are as in 
    \eqref{delta-min-optimal-even-max} if $d$ is even,
    while 
    $\delta_j^\ast$ are as in 
    \eqref{delta-min-optimal-odd-max} if $d$ is odd.
This give the desired solution to \eqref{LP-symmetric} with $\max$ instead of $\min$,
    and proves Theorem \ref{sol:max-value}.


\section{maximal negative volume problem in dimensions $d\leq 6$ analytically}
\label{sec:remaining-cases}

The solutions to the maximal negative volume quasi-copula problem in dimensions $2\leq d\leq 6$ are already known, i.e., see \cite{Nelsen2002} for $d=2$, \cite{BMUF07} for $d=3$, \cite{UF23} for $d=4$ and \cite{BeOmVuZa}
for $d=5,6$.
However, the solutions for $d\geq 3$ are numerical, based on the solution to the corresponding linear programs using computer software. In this section we will present analytical solutions. \\

Assume the notation as in Section \ref{symmetrization}.
We will solve the linear program
\eqref{dual-LP-symmetric-v4-new} and consequently \eqref{LP-symmetric}
analytically for $3\leq d\leq 6$.

Let $(l^{\opt}_1,\ldots,l^{\opt}_d,y^{\opt}_3)$ be an optimal solution
to \eqref{dual-LP-symmetric-v4-new}.
By the moreover part of Claim 3 in the proof of Proposition \ref{prop:simplification}, we have {to analyse the possibilities} $l_d=0$ for $d\leq 6$. {If any of these gives a feasible solution, then there also exists an optimal solution with $l_d^{\ast}=0$, as $y_3$ is at most $1$ in this case. Below we derive the optimal solutions to these programs for $l_d=0$. To complete the solution one needs to find a realization of the $d$-quasi-copula with such $d$-box but these are already given in the references from the first paragraph above.}


\subsection{Case $d=3$}
The linear program \eqref{dual-LP-symmetric-v4-new} for $d=3$ is

\begin{align*}
\begin{split}
-\Big(\min_{
\substack{
    l_1,l_2,
    y_3
}}
&\hspace{0.5cm} 2y_3\Big),\\[0.3em]
\text{subject to }
&\hspace{0.5cm}
    3y_3 = l_1+l_2,\\
&\hspace{0.5cm}
    l_1+ 2y_3\ge -\binom{2}{0}=-1\\[0.3em]
&\hspace{0.5cm}
    l_2+ 2y_3\ge \binom{2}{1}=2\\[0.3em]
&\hspace{0.5cm} 
    1\geq y_3\geq 0,\; 
    l_1\geq 0,\; 
    l_2\geq 0.
\end{split}
\end{align*}
Repeating the arguments from the proof of Proposition \ref{auxiliary-proposition},
in the optimal solution $(l_1^\ast,l_2^\ast,y_3^\ast)$ we have that 
$l_1^\ast=0$ and $l_2^\ast+2y_3^\ast=2$. Using this in
$3y^\ast_3 = l_1^\ast+l_2^\ast$ we get $y_3^\ast=\frac{2}{5}$.
So the minimal volume box of some $3$-quasi-copula has value $-\frac{4}{5}$. 
Hence, an optimal solution 
$(y_1^\ast,l_1^\ast,l_2^\ast,l_3^\ast,y_2^\ast,y_3^\ast)$
to \eqref{dual-LP-symmetric-v3} for $d=3$
is 
$\big(0,0,\frac{2}{5},0,\frac{6}{5},\frac{2}{5}\big)$.
Using the complementary slackness, an optimal solution 
$(a^\ast,b^\ast,q_0^\ast,\delta_1^\ast,\delta_2^\ast,\delta_3^\ast)$
to its dual \eqref{LP-symmetric-v3} 
is equal to
$\big(\frac{2}{5},\frac{4}{5},0,0,\frac{2}{5},0\big).$
Finally, an optimal solution 
to \eqref{LP-symmetric} is 
$$
(a^\ast,b^\ast,q_0^\ast,q_1^\ast,q_2^\ast,q_3^\ast)
= \Big(\frac{2}{5},\frac{4}{5},0,0,\frac{2}{5},\frac{2}{5}\Big).
$$


\subsection{Case $d=4$}
The linear program \eqref{dual-LP-symmetric-v4-new} for $d=4$ is

\begin{align*}
\begin{split}
-\Big(\min_{
\substack{
    l_1,l_2,l_3,
    y_3
}}
&\hspace{0.5cm} 3y_3\Big),\\[0.3em]
\text{subject to }
&\hspace{0.5cm}
    4y_3 = l_1+l_2+l_3,\\
&\hspace{0.5cm}
    l_1+ 3y_3\ge \binom{3}{0}=1\\[0.3em]
&\hspace{0.5cm}
    l_2+ 3y_3\ge -\binom{3}{1}=-3\\[0.3em]
&\hspace{0.5cm}
    l_3+ 3y_3\ge \binom{3}{2}=3\\[0.3em]
&\hspace{0.5cm} 
    1\geq y_3\geq 0,\; 
    l_1\geq 0,\; 
    l_2\geq 0,\;
    l_3\geq 0.
\end{split}
\end{align*}
Repeating the arguments from the proof of Proposition \ref{auxiliary-proposition},
in the optimal solution $(l_1^\ast,l_2^\ast,y_3^\ast)$ we have that 
$l_2^\ast=0$ and one of the cases:
\begin{itemize}
    \item $l_1^\ast+3y_3^\ast=1$ and $l_3^\ast+3y_3^\ast=3$, or
    \smallskip
    \item $l_1^\ast=0$ and $l_3^\ast+3y_3^\ast=3$,
\end{itemize} 
Using these in $4y^\ast_3 = l_1^\ast+l_2^\ast$ we get $y_3^\ast=\frac{2}{5}$
in the first case and $y_3^\ast=\frac{3}{7}$ in the second case.
In the first case $l_1^\ast=-\frac{1}{5}<0$, which is not in the feasible region.
So the second case applies and the minimal volume box of some $4$-quasi-copula has value $-\frac{9}{7}$. 
Hence, an optimal solution 
$(y_1^\ast,l_1^\ast,l_2^\ast,l_3^\ast,y_2^\ast,y_3^\ast)$
to \eqref{dual-LP-symmetric-v3} for $d=4$
is 
$\big(0,0,0,\frac{12}{7},\frac{12}{7},\frac{3}{7}\big)$.
Using the complementary slackness, an optimal solution 
$(a^\ast,b^\ast,q_0^\ast,\delta_1^\ast,\delta_2^\ast,\delta_3^\ast,\delta_4^\ast)$
to its dual \eqref{LP-symmetric-v3} 
is equal to
$\big(\frac{3}{7},\frac{6}{7},0,0,0,\frac{3}{7},0\big).$
Finally,  an optimal solution 
to \eqref{LP-symmetric} is 
$$
    (a^\ast,b^\ast,q_0^\ast,q_1^\ast,q_2^\ast,q_3^\ast,q_4^\ast)
    =\Big(\frac{3}{7},\frac{6}{7},0,0,0,\frac{3}{7},\frac{3}{7}\Big).
$$


\subsection{Case $d=5$}
The linear program \eqref{dual-LP-symmetric-v4-new} for $d=5$ is

\begin{align*}
\begin{split}
-\Big(\min_{
\substack{
    l_1,l_2,l_3,l_4,
    y_3
}}
&\hspace{0.5cm} 4y_3\Big),\\[0.3em]
\text{subject to }
&\hspace{0.5cm}
    5y_3 = l_1+l_2+l_3+l_4,\\
&\hspace{0.5cm}
    l_1+ 4y_3\ge -\binom{4}{0}=-1\\[0.3em]
&\hspace{0.5cm}
    l_2+ 4y_3\ge \binom{4}{1}=4\\[0.3em]
&\hspace{0.5cm}
    l_3+ 4y_3\ge -\binom{4}{2}=-6\\[0.3em]
&\hspace{0.5cm}
    l_4+ 4y_3\ge \binom{4}{3}=4\\[0.3em]
&\hspace{0.5cm} 
    1\geq y_3\geq 0,\; 
    l_1\geq 0,\; 
    l_2\geq 0,\;
    l_3\geq 0,\;
    l_4\geq 0.
\end{split}
\end{align*}
Repeating the arguments from the proof of Proposition \ref{auxiliary-proposition},
in the optimal solution $(l_1^\ast,l_2^\ast,l_3^\ast,l_4^\ast,y_3^\ast)$ we have that 
$l_1^\ast=l_3^\ast=0$ and $l_2^\ast+4y^\ast_3=l_4^\ast+4y_3^\ast=4$. 
Using these in $5y^\ast_3 = l_2^\ast+l_4^\ast$ we get $y_3^\ast=\frac{8}{13}$
So the minimal volume box of some $5$-quasi-copula has value $-\frac{32}{13}$. 
Hence, an optimal solution 
$(y_1^\ast,l_1^\ast,l_2^\ast,l_3^\ast,l_4^\ast.y_2^\ast,y_3^\ast)$
to \eqref{dual-LP-symmetric-v3} for $d=5$
is 
$\big(0,0,\frac{20}{13},0,\frac{20}{13},\frac{30}{13},\frac{8}{13}\big)$.
Using the complementary slackness, an optimal solution 
$(a^\ast,b^\ast,q_0^\ast,\delta_1^\ast,\delta_2^\ast,\delta_3^\ast,\delta_4^\ast,\delta_5^\ast)$
to its dual \eqref{LP-symmetric-v3} 
is equal to
$\big(\frac{8}{13},\frac{12}{13},0,0,\frac{4}{13},0,\frac{4}{13},0\big).$
Finally,  an optimal solution 
to \eqref{LP-symmetric} is 
$$
    (a^\ast,b^\ast,q_0^\ast,q_1^\ast,q_2^\ast,q_3^\ast,q_4^\ast,q_5^\ast)
    =\Big(\frac{8}{13},\frac{12}{13},0,0,\frac{4}{13},\frac{4}{13},\frac{8}{13},\frac{8}{13}\Big).
$$


\subsection{Case $d=6$}
The linear program \eqref{dual-LP-symmetric-v4-new} for $d=6$ is

\begin{align*}
\begin{split}
-\Big(\min_{
\substack{
    l_1,l_2,l_3,l_4,l_5,
    y_3
}}
&\hspace{0.5cm} 5y_3\Big),\\
\text{subject to }
&\hspace{0.5cm}
    6y_3 = l_1+l_2+l_3+l_4+l_5+l_6,\\
&\hspace{0.5cm}
    l_1+ 5y_3\ge \binom{5}{0}=1\\[0.3em]
&\hspace{0.5cm}
    l_2+ 5y_3\ge -\binom{5}{1}=-5\\[0.3em]
&\hspace{0.5cm}
    l_3+ 5y_3\ge \binom{5}{2}=10\\[0.3em]
&\hspace{0.5cm}
    l_4+ 5y_3\ge -\binom{5}{3}=-10\\[0.3em]
&\hspace{0.5cm}
    l_5+ 5y_3\ge \binom{5}{4}=5\\[0.3em]
&\hspace{0.5cm} 
    1\geq y_3\geq 0,\; 
    l_1\geq 0,\; 
    l_2\geq 0,\;
    l_3\geq 0,\;
    l_4\geq 0,\;
    l_5\geq 0.
\end{split}
\end{align*}
Repeating the arguments from the proof of Proposition \ref{auxiliary-proposition},
in the optimal solution $(l_1^\ast,l_2^\ast,l_3^\ast,l_4^\ast,l_5^\ast,y_3^\ast)$ we have that 
$l_2^\ast=l_4^\ast=0$ and 
 and one of the cases:
\begin{itemize}
    \item 
        $l_1^\ast+5y_3^\ast=1$, 
        $l_3^\ast+5y_3^\ast=10$ and 
        $l_5^\ast+5y_3^\ast=5$, or
    \smallskip
    \item 
        $l_1^\ast=0$,
        $l_3^\ast+5y_3^\ast=10$ and 
        $l_5^\ast+5y_3^\ast=5$, or
    \smallskip
    \item 
        $l_1^\ast=0$,
        $l_3^\ast+5y_3^\ast=10$ and 
        $l_5^\ast=0$. 
\end{itemize} 
Using these in $6y^\ast_3 = l_1^\ast+l_3^\ast+l_5^\ast$ 
we get 
    $y_3^\ast=\frac{16}{21}$
in the first case,
    $y_3^\ast=\frac{15}{16}$ 
in the second case
and
    $y_3^\ast=\frac{10}{11}$ 
in the third case.
In the smallest candidate for the solution, i.e., $\frac{16}{21}$,
we have that $l_1^\ast=-\frac{59}{21}<0$, which is not in the feasible region.
The second smallest candidate for the solution, i.e., $\frac{15}{16}$
indeed comes from the feasible (and hence optimal) solution, i.e.,
$
(l_1^\ast,l_2^\ast,l_3^\ast,l_4^\ast,l_5^\ast,y_3^\ast)
=
\big(0,0,\frac{85}{16},0,\frac{5}{16},\frac{15}{16}\big).
$
So the minimal volume box of some $6$-quasi-copula has value $-\frac{75}{16}$. 
Hence, an optimal solution 
$(y_1^\ast,l_1^\ast,l_2^\ast,l_3^\ast,l_4^\ast,l_5^\ast,l_6^\ast,y_2^\ast,y_3^\ast)$
to \eqref{dual-LP-symmetric-v3} for $d=6$
is 
$\big(0,0,0,\frac{85}{16},0,\frac{5}{16},0,\frac{90}{16},\frac{15}{16}\big)$.
Using the complementary slackness, an optimal solution 
$(a^\ast,b^\ast,q_0^\ast,\delta_1^\ast,\delta_2^\ast,\delta_3^\ast,\delta_4^\ast,\delta_5^\ast,\delta_6^\ast)$
to its dual \eqref{LP-symmetric-v3} 
is equal to
$\big(
\frac{5}{8},\frac{15}{16},0,0,0,\frac{5}{16},0,\frac{5}{16},0
\big).$
Finally,  an optimal solution 
to \eqref{LP-symmetric} is 
$$
    (a^\ast,b^\ast,q_0^\ast,q_1^\ast,q_2^\ast,q_3^\ast,q_4^\ast,q_5^\ast,q_6^\ast)
    =\Big(\frac{5}{8},\frac{15}{16},0,0,0,\frac{5}{16},\frac{5}{16},
    \frac{5}{8},\frac{5}{8}\Big).
$$

\bigskip


{
\section{Concluding remarks and future research}
In this paper, we addressed the problem of extreme values of the mass distribution associated with a multidimensional quasi-copula, following the approach of reformulating the problem in the linear programming language introduced in \cite{BMUF07} for the three-dimensional case.
As shown in our previous work \cite{BeOmVuZa}, solving the linear program using computer software becomes infeasible already at relatively low dimensions (less than 20). However, the symmetry of the solutions suggested that this symmetry could also be preserved in a general dimension. This was a crucial observation that allowed us to simplify the problem by symmetrizing the linear program (see \eqref{LP-symmetric} and
\eqref{LP-symmetric-v3})
and then solving it explicitly via the corresponding dual linear program (see \eqref{dual-LP-symmetric-v4-new} and \eqref{dual-LP-symmetric-v4-max-new}). Moreover, with this approach we can also explain why dimensions up to six are special for the maximal negative volume question (see Remark \ref{remark-after-theorems}.\eqref{remark-after-theorems-pt2}).
However, it turns out that a closed form solution for the extreme values does not exist, which is also supported by the numerical analysis of the solutions in Section \ref{sec:numerical-aspects}.\\

Finally, we give some directions for future research.
To explain the difference in extreme volumes between quasi-copulas and copulas, it would be interesting to study a larger class of aggregation functions (see \cite[Chapter 8]{durante2015principles}, \cite{ARIASGARCIA20201,AGMB17})
such as (Lipschitz-continuous or continuous) semi-copulas on the one hand and intermediate classes of $k$-dimensional increasing $d$-quasi-copulas for $k\in \{2,\ldots,d-1\}$ on the other hand.
Note that $k=1$ and $k=d$ are exactly quasi-copulas and copulas, respectively. Understanding the behaviour of the extreme volumes for these classes would show how important is the size of $k$ relative to $d$ and thus explain why copulas are so much more restrictive than quasi-copulas in terms of volumes. 

It would be also interesting to characterize extreme points and faces of the convex set of all $d$-quasi-copulas, since by Krein-Milman theorem every quasi-copula is a convex combinations of extreme points of the set of quasi-copulas. Clearly, extreme points form a subset of the extreme volume points, but the other inclusion does not necessarily hold. The symmetric quasi-copulas from Theorem \ref{sol:min-value} and \ref{sol:max-value} are examples of extreme volume solutions which are not necessarily extreme points, since they are obtained as convex combinations of non-symmetric extreme volume solutions (see \eqref{average}).
}

\bigskip

 \noindent \textbf{Acknowledgement.}\
We would like to thank anonymous reviewers for carefully reading our manuscript and suggestions to improve the presentation of our results.


\begin{thebibliography}{99}

\bibitem{Als93}
C. Alsina,  R.B. Nelsen and B. Schweizer,  
  On the characterization of a class of binary operations on distribution functions,
  \textit{Statistics and Probability Letters}
  17 (1993) 85--89.

{
\bibitem{AGMB17}
J.J. Arias-García, R. Mesiar and B. {De Baets},
The unwalked path between quasi-copulas and copulas: Stepping stones in higher dimensions,
\textit{International Journal of Approximate Reasoning},
80 (2017) 89--99.
}

\bibitem{ARIASGARCIA20201}
J.J. Arias-García, R. Mesiar and B. {De Baets},
A hitchhiker's guide to quasi-copulas,
\textit{Fuzzy Sets and Systems}
393 (2020), 1--28.


\bibitem{2006217}
B. {De Baets}, H. {De Meyer}, B. De Schuymer and S. Jenei
{Cyclic evaluation of transitivity of reciprocal relations},
\textit{Social Choice and Welfare},
26 (2006), 217--238.

\bibitem{DEBAETS20061463}
{B. {De Baets}, S. Janssens and H. {De Meyer}},
{Meta-theorems on inequalities for scalar fuzzy set cardinalities},
\textit{Fuzzy Sets and Systems},
157 (2006) {1463--1476}.

\bibitem{BMUF07}
B. {De Baets}, H. {De Meyer} and M. {\'U}beda-Flores,
{Extremes of the mass distribution associated with a trivariate quasi-copula},
\textit{Comptes Rendus Mathematique},
344 (2007), 587--590.

\bibitem{BeOmVuZa}
M. Bel\v sak, M. Omladi\v c, M. Vuk, A. Zalar,
{Extreme values of the mass distribution associated with $d$-quasi-copulas via linear programming}, 	\textit{Fuzzy Sets and Systems},
517 (2025) 17 pp.

\bibitem{bezanson_julia_2017}
J. Bezanson, A. Edelman, S. KarpinskI V.B Shah,
	{Julia: A Fresh Approach to Numerical Computing},
 \textit{SIAM Rev.}, 59 (2017), 65--98.
 
\bibitem{Cuculescu2001731}
	{I. Cuculescu and R. Theodorescu},
	{Copulas: Diagonals and tracks},
	\textit{Rev. Roumaine Math. Pures Appl.},
	{46} (2001), 731--742.

\bibitem{durante2015principles}
F. Durante and C. Sempi,
  \textit{Principles of Copula Theory},
  CRC Press, 2015.

\bibitem{FERNANDEZSANCHEZ20111365}
J. Fernández-Sánchez, R.B. Nelsen and M. {\'U}beda-Flores,
{Multivariate copulas, quasi-copulas and lattices},
\textit{Statistics and Probability Letters},
81 (2011), {1365--1369}.

\bibitem{FERNANDEZSANCHEZ2014109}
J. Fernández-Sánchez and M. {\'U}beda-Flores,
{A note on quasi-copulas and signed measures},
\textit{Fuzzy Sets and Systems},
234 (2014) 109--112.

\bibitem{Gen99}
C. Genest,  J.J. Quesada Molina, J.A. Rodr{\'i}guez Lallena and C. Sempi,
  {A Characterization of Quasi-copulas},
  \textit{Journal of Multivariate Analysis}
  69 (1999), 193--205.


\bibitem{Highs}
Q. Huangfu and J.A.J. Hall,
{Parallelizing the dual revised simplex method},
\textit{Mathematical Programming Computation}
10 (2018) 119--142.

\bibitem{Klemetal} E.P.~Klement, D.~Kokol Bukov\v{s}ek, M.~Omladi\v{c}, S.~Saminger-Platz, N.~Stopar, {Multivariate copulas with given values at two arbitrary points}, \textit{Stat.\ Papers} \textbf{64} (2023), 2015--2055.


\bibitem{KOLESAROVA20121}
A. Kolesárová, A. Stupňanová and J. Beganová,
{Aggregation-based extensions of fuzzy measures},
\textit{Fuzzy Sets and Systems}
194 (2012) 1--14.

\bibitem{Nel05}
R.B. Nelsen and M. {\'U}beda-Flores,
  The lattice-theoretic structure of sets of bivariate copulas and quasi-copulas,
  \textit{Comptes Rendus. Math{\'e}matique}
  341 (2005) 583--586.

\bibitem{Nelsen2002}
R.B. Nelsen, J.J. Quesada-Molina, J.A. Rodr{í}guez-Lallena and
M. {\'U}beda-Flores,
{Some New Properties of Quasi-Copulas},
\textit{Distributions With Given Marginals and Statistical Modelling},
Springer Netherlands,
(2002), 187--194.

\bibitem{NQMRLUF02}
R.B. Nelsen, J.J. Quesada-Molina, J.A. Rodr{í}guez-Lallena and
M. {\'U}beda-Flores,
  {Some new properties of quasi-copulas},
  \textit{Distributions with given marginals and Statistical Modelling} (2002), 187--194.

\bibitem{NELSEN2005583}
R.B. Nelsen and M. {{\'U}beda Flores},
{The lattice-theoretic structure of sets of bivariate copulas and quasi-copulas},
\textit{Comptes Rendus Mathematique},
341 (2005) 583--586.


\bibitem{OmSt1} M. Omladi\v{c}, N. Stopar, {Final solution to the problem of relating a true copula to an imprecise copula}, 
\textit{Fuzzy Sets and Systems} \textbf{393} (2020), 96--112.

\bibitem{OmSt5} M.~Omladi\v c, N.~Stopar, {Dedekind-MacNeille completion of multivariate copulas via ALGEN method}, \textit{Fuzzy Sets and Systems} \textbf{441} (2022), 321--334.


\bibitem{RODRIGUEZLALLENA2009717}
J.A. Rodríguez-Lallena and M. {\'U}beda-Flores,
{Some new characterizations and properties of quasi-copulas},
\textit{Fuzzy Sets and Systems}, 160 (2009), 717--725.

\bibitem{Sklar1959229}
A. Sklar, Fonctions de r\'epartition \`a ndimensions et leurs marges, 
    \textit{Publ. Inst. Stat. Univ. Paris},
    8 (1959), 229--231.

\bibitem{St} 
N. Stopar, {Bivariate measure-inducing quasi-copulas}, \textit{Fuzzy Sets and Systems} \textbf{492} (2024), 109068.

\bibitem{UF23}
M. {\'U}beda-Flores, Extreme values of the mass distribution associated with a tetravariate quasi-copula,
\textit{Fuzzy Sets and Systems}
473 (2023).

\bibitem{Quasi-copula-Gitlab}
M. Vuk,
\textit{QuasiCopula.jl},
\texttt{https://gitlab.com/mrcinv/QuasiCopula.jl}, 
Accessed: 15 March 2025.

\bibitem{ram2024}
	Wolfram Research, Inc., Mathematica, Version 14.1, Wolfram Research, Inc., Champaign, IL, 2024.


\end{thebibliography}
 \end{document}